\documentclass[conference]{IEEEtran}
\pdfoutput=1
\usepackage{times}

\usepackage[numbers]{natbib}
\usepackage{multicol}
\usepackage[hidelinks,bookmarks=true]{hyperref}
\usepackage{amsmath,amssymb,amsfonts}
\usepackage{xspace, xcolor}
\usepackage{graphicx}
\usepackage{algorithm}
\usepackage{algorithmicx}
\usepackage[noend]{algpseudocode}
    \algrenewcommand\algorithmicindent{0.75em}%
\usepackage[textsize=scriptsize,textwidth=1cm]{todonotes}
\usepackage[]{subcaption}
\usepackage{tikz}
\usetikzlibrary{fit,calc}
\usepackage{wrapfig}
\graphicspath{{figures/}}
\hypersetup{bookmarksopen,bookmarksnumbered,
colorlinks=false
}

\newcommand{\ignore}[1]{}

 \def\I{\mathcal{I}} \def\E{\mathcal{E}}
\def\S{\mathcal{S}} \def\G{\mathcal{G}} 
\def\I{\mathcal{I}} \def\T{\mathcal{T}} 
 \def\V{\mathcal{V}} 
 \def\W{\mathcal{W}} 
 \def\X{\mathcal{X}} 
 
\def\R{\mathcal{I}}

\def\eps{\varepsilon}

\newcommand{\Cpp}{C\raise.08ex\hbox{\tt ++}\xspace}

\newcommand{\PSPACE}{{\small \ensuremath{\mathsf{PSPACE}}\xspace}}
\newcommand{\NP}{{\small \ensuremath{\mathsf{NP}}\xspace}}

\newtheorem{cor}{Corollary}

\newtheorem{defin}{Definition}

\newcommand\algname[1]{\textsf{#1}\xspace}
\newcommand\mhastar{\algname{MHA*}}

\newcommand\astar{\algname{A*}}
\newcommand\rrt{\algname{RRT}}
\newcommand\rrg{\algname{RRG}}
\newcommand\rrtot{\algname{RRTOT}}
\newcommand\rita{\algname{RITA}}
\newcommand\prm{\algname{PRM}}
\newcommand\iris{\algname{IRIS}}

\newcommand\pap{\algname{PAP}}
\newcommand\paps{\algname{PAP}'s\xspace}
\newcommand\pp{\algname{PP}}
\newcommand\pps{\algname{PP}'s\xspace}

\def\naive{{na\"{\i}ve}\xspace}

\colorlet{pink}{red!40}
\colorlet{light_blue}{cyan!60}

\newcommand{\Lim}[1]{\raisebox{0.5ex}{\scalebox{0.8}{$\displaystyle \lim_{#1}\;$}}}

\begin{document}

\title{Toward Asymptotically-Optimal Inspection Planning
via Efficient Near-Optimal Graph Search}

\author{\authorblockN{Mengyu Fu\authorrefmark{1},
Alan Kuntz\authorrefmark{1},
Oren Salzman\authorrefmark{2}, and
Ron Alterovitz\authorrefmark{1}}
\authorblockA{\authorrefmark{1}Department of Computer Science, University of North Carolina at Chapel Hill, Chapel Hill, NC 27599, USA \\ 
Email: {\tt\small \{mfu,adkuntz,ron\}@cs.unc.edu}}
\authorblockA{\authorrefmark{2}Robotics Institute, Carnegie Mellon University, Pittsburgh, PA 15213, USA\\
Email: {\tt\small osalzman@andrew.cmu.edu}}}

\maketitle

\begin{abstract}
Inspection planning, the task of planning motions that allow a robot to inspect a set of points of interest, has applications in domains such as industrial, field, and medical robotics.
Inspection planning can be computationally challenging, as the search space over motion plans grows exponentially with the number of points of interest to inspect.
We propose a novel method, Incremental Random Inspection-roadmap Search (\iris), 
that computes inspection plans whose length and set of successfully inspected points asymptotically converge to those of an optimal inspection plan.
\iris incrementally densifies a motion-planning roadmap using a sampling-based algorithm, and performs efficient near-optimal graph search over the resulting roadmap as it is generated.
We  demonstrate \iris's efficacy on a simulated planar 5DOF manipulator inspection task and on a medical endoscopic inspection task for a continuum parallel surgical robot in cluttered anatomy segmented from patient CT data.
We show that \iris computes higher-quality inspection plans orders of magnitudes faster than a prior state-of-the-art method.
\end{abstract}

\IEEEpeerreviewmaketitle

\section{Introduction}
In this work we investigate the problem of \emph{inspection planning}, or coverage planning~\cite{almadhoun2016survey,galceran2013RAS}.
Here, we consider the specific setting where we are given a robot equipped with a sensor and a set of points of interest (POI) in the environment to be inspected by the sensor.
The problem calls for computing a minimal-length motion plan for the robot that maximizes the number of POI inspected.
This problem has a multitude of diverse applications, including
industrial surface inspections in production lines~\cite{RMGM13}, 
surveying the ocean floor by autonomous underwater vehicles~\cite{bingham2010robotic, gracias2013mapping, johnson2010generation, tivey1997autonomous}, structural inspection of bridges using aerial robots~\cite{bircher2015structural,bircher2016three},
and medical applications such as inspecting patient anatomy during surgical procedures~\cite{Kuntz2018KinematicDO}, which motivates this work.

\begin{figure}[t!]
  {\includegraphics[width = \linewidth]{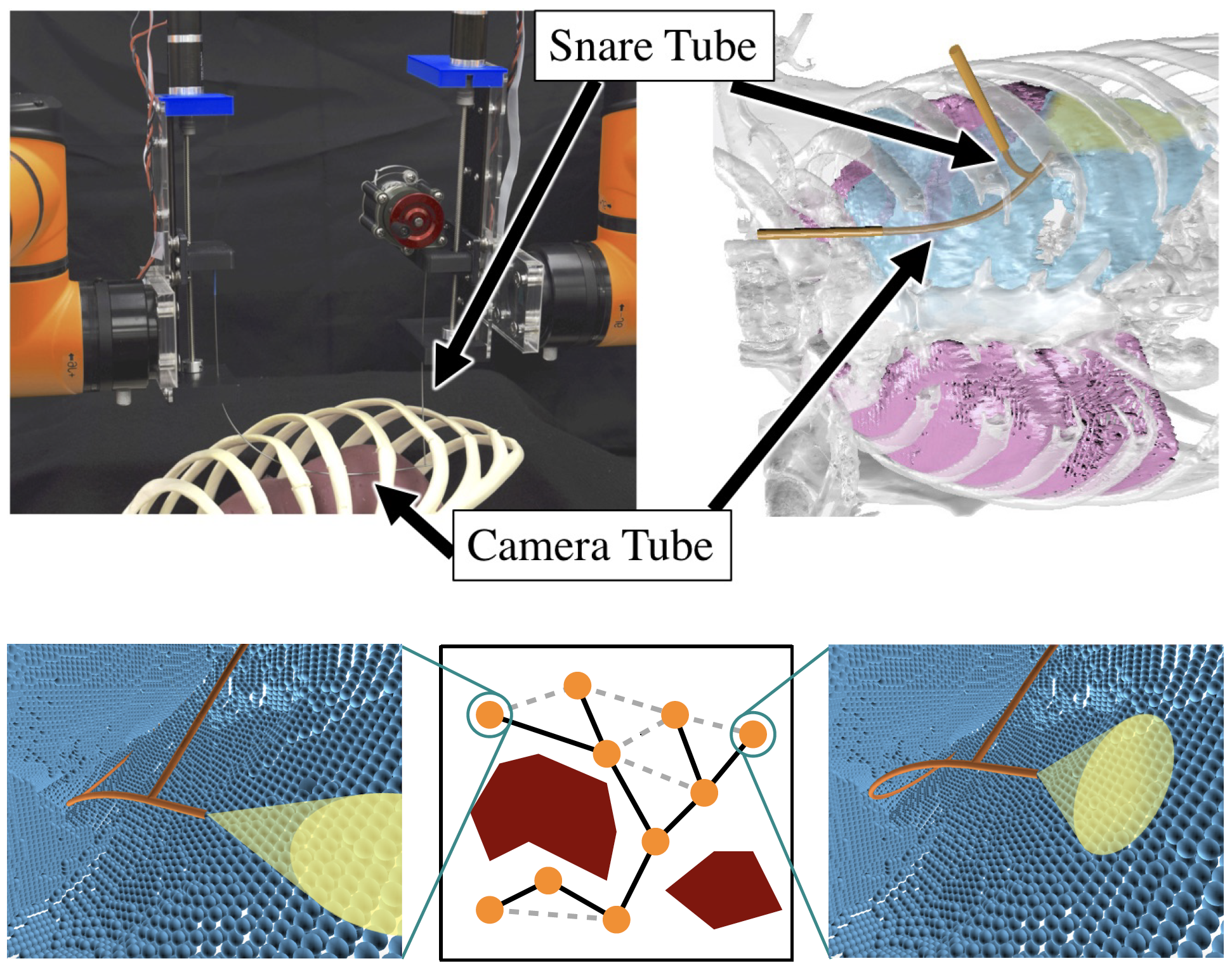}}
  \caption{Inspection planning in human anatomy.
    \protect Top~Left: The Continuum Reconfigurable Incisionless Surgical Parallel (CRISP) robot~\cite{AMW17,MASMW16} is composed of needle-diameter tubes assembled into a parallel structure inside the patient's body (in which a tube uses a snare system to grip a tube with a camera affixed to its tip) and then robotically manipulated outside the body, allowing for smaller incisions and faster recovery times compared to traditional endoscopic tools (which have larger diameters).
    \protect Top~Right: The CRISP robot in simulation inspecting a collapsed lung, a scenario segmented from a CT scan of a real patient with this condition.  The visualization shows the robot (orange), the lungs (pink), and the pleural surface visible (green) and not visible (blue) by the robot's camera sensor in its current configuration.
    \protect Bottom:~Our method, \iris, constructs a tree representing collision-free configurations (orange nodes) and motions (solid lines) of the robot. For each node, the robot (orange) can see points of interest (yellow) on the segmented anatomy (blue). \iris then searches over an implicit graph structure (dashed lines) to compute asymptotically-optimal inspection plans.}
 \label{fig:inspection}
\end{figure}

Na\"ively-computed inspection plans may enable inspection of only a subset of the POI and may require motion plans orders of magnitude longer than an optimal plan, and hence may be undesirable or infeasible due to battery or time constraints.
In medical applications, physicians may want to maximize the number of POI inspected for diagnostic purposes.
Additionally, the procedure should be completed as fast as is safely possible to reduce costs and improve patient outcomes, especially if the patient is under anesthesia during the procedure.
For example, a robot assisting in the diagnosis of the cause of a pleural effusion (a serious medical condition which causes the collapse of a patient's lung) will need to visually inspect the surface of the collapsed lung and chest wall inside the body in as short a time as possible (see Fig.~\ref{fig:inspection}).
We note that it may not be possible to inspect some POI due to obstacles in the anatomy and the kinematic constraints of the robot. 
Our goal is to compute kinematically-feasible collision free inspection plans that maximizes the number of POI inspected, and of the motion plans that inspect those POI we compute a shortest plan.

Inspection planning is computationally challenging because the search space is embedded in a high-dimensional \emph{configuration space}~$\X$ 
(space of all parameters that determine the shape of the robot)~\cite{CBHKKLT05,Lat91,LaValle2006_Book}.
Even finding the shortest plan between two points in~$\X$ that avoid obstacles (without reasoning about inspection) is computationally hard.\footnote{Computing the shortest plan for a point robot moving amidst polyhedral obstacles in 3D is \NP-hard, and many variants of the general motion planning problem are \PSPACE-hard. For further details, see~\cite{HSS17b}.} 
If we want a minimum-length motion plan that maximizes the number of POI inspected, our problem is accentuated as we have to simultaneously reason about the system's constraints, motion plan length, and POI inspected.

There are multiple approaches to computing inspection plans.
Optimization-based methods locally search over the space of all inspection plans~\cite{bircher2015structural,bogaerts2018gradient}.
Decoupled approaches first independently select suitable viewpoints and then determine a visiting sequence, i.e., a motion plan for the robot that realizes this sequence~\cite{DK00,EH11}.
Finally, recent progress in motion planning~\cite{KF11} has enabled methods to exhaustively search over the space of all motion plans~\cite{BASOBS17,kafka2016random,papadopoulos2013asymptotically} thus guaranteeing asymptotic optimality, a key requirement in many applications, including medical ones.
Roughly speaking, asymptotic optimality for inspection planning means these methods produce inspection plans whose length and number of points inspected will asymptotically converge to those of an optimal inspection plan, given enough planning time.

Of all the aforementioned methods, only algorithms in the latter group provide any formal guarantees on the quality of the solution.
This guarantee is achieved by attempting to exhaustively compute the set of Pareto-optimal inspection plans
embedded in~$\X$.
In our setting, the set of Pareto-optimal inspection plans is the minimal set of inspection plans such that each plan is either shorter or has better coverage of the POI than any other inspection plan\footnote{More formally, a plan~$P$ connecting two configurations $\mathbf{q},\mathbf{q}'\in\X$ is said to be Pareto optimal in our setting if any other plan connecting $\mathbf{q}$ to $\mathbf{q}'$ is either longer or does not inspect a point visible to $P$.}.
Unfortunately, this comes at the price of very long computation times as the size of the search space is exponential in the number of POI.

\begin{figure}[tb]
  \centering
 	\includegraphics[height=4.5cm]{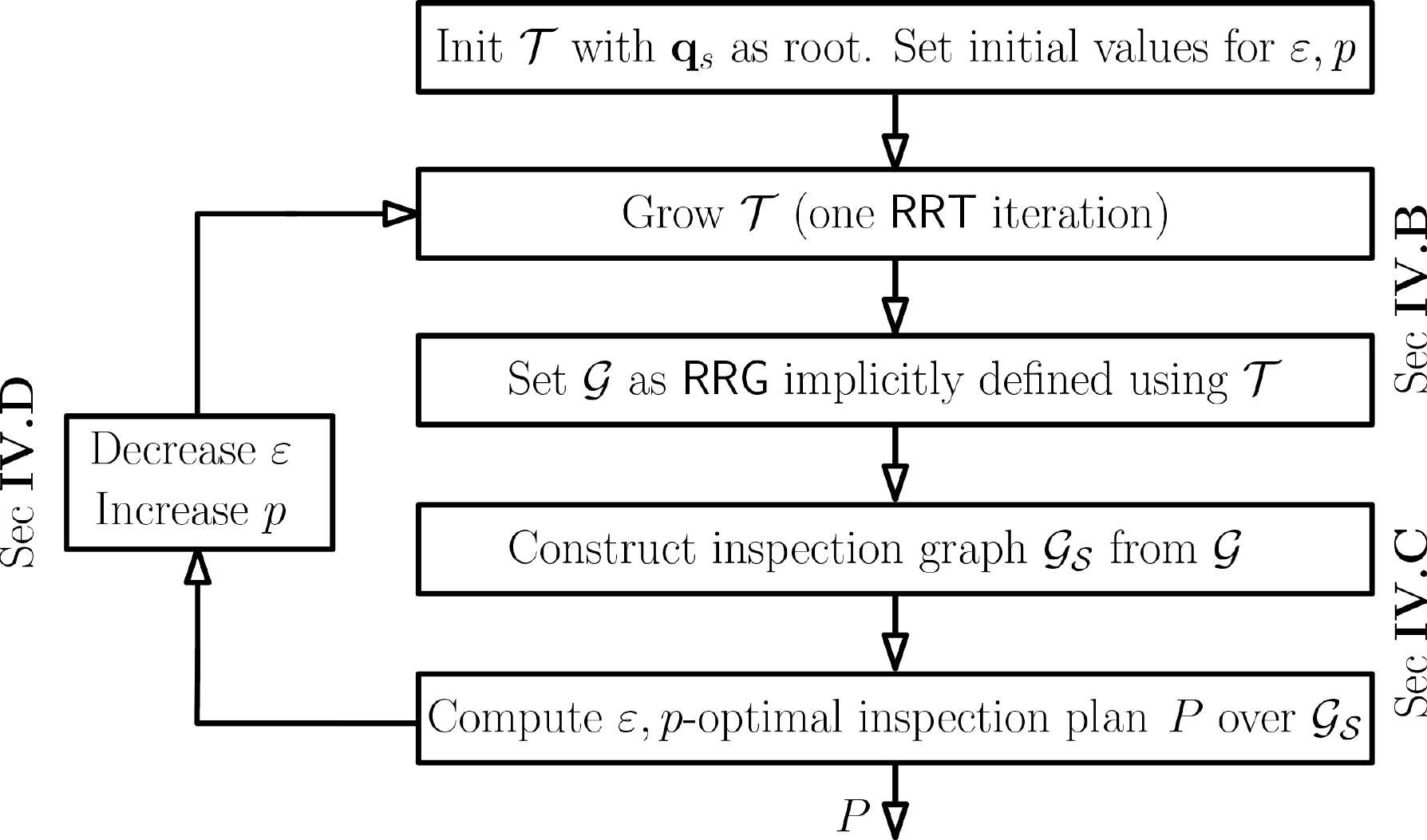}
  \caption{Overview of the \iris algorithmic framework}
 \label{fig:method}
\end{figure}

To this end, we introduce Incremental Random Inspection-roadmap Search (\iris), a new asymptotically-optimal inspection-planning algorithm.
\iris incrementally constructs a sequence of increasingly dense roadmaps---graphs embedded in~$\X$ wherein each vertex represents a collision-free configuration and each edge a collision-free transition between configurations---and computes an inspection plan on the roadmaps as they are constructed (see Fig.~\ref{fig:method}).
Unfortunately, even the problem of computing an optimal inspection plan on a graph (and not in the continuous space) is computationally hard.
To this end, our key insight is to compute a \emph{near-optimal} inspection plan on each roadmap. 
Namely, we compute an inspection plan that is at most $1+\eps$ the length of an optimal plan while covering at least $p$-percent of the number of POI (for any $\eps \geq 0$ and $p \in (0,1]$).
This additional flexibility allows us to improve the quality of our inspection plan in an \emph{anytime} manner, i.e., the algorithm can be stopped at any time and return the best inspection plan found up until that point.
We achieve this by incrementally densifying the roadmap and then searching over the densified roadmap for a near-optimal inspection plan---a process that is repeated as time allows.
By reducing the approximation factor between iterations, we ensure  that our method is asymptotically optimal.

The key contribution of our work is a computationally-efficient algorithm to compute provably near-optimal inspection plans on graphs. 
By pruning away large portions of the search space, this algorithmic building block enables us to dramatically outperform Rapidly-exploring Random Tree Of Trees (\rrtot)~\cite{BASOBS17}---a state-of-the-art asymptotically-optimal inspection planner.
Specifically, we demonstrate the efficacy of our approach in simulation for a continuum robot with complex dynamics---the needle-diameter Continuum Reconfigurable Incisionless Surgical Parallel (CRISP) robot~\cite{AMW17,MASMW16}, working in a medically-inspired setting involving diagnosis of a pleural effusion (see Fig.~\ref{fig:inspection}).

\section{Related Work}

\subsection{Sampling-based motion planning}
\label{sec:sbmp}
Motion planning algorithms aim to compute a collision-free motion for a
robot to accomplish a task in an environment cluttered with obstacles~\cite{HSS17b,LaValle2006_Book, LP17}.
A common approach to motion planning is by \emph{sampling-based} algorithms that construct a \emph{roadmap}. 
Examples include the Probabilistic Roadmaps ({\prm}s)~\cite{Kavraki1996_TRA} (for solving multiple queries) and the  Rapidly-exploring Random Trees ({\rrt}s)~\cite{LaValle2001_IJRR} for solving single queries.
These methods, and many variations thereof, are \emph{probabilistically complete}---namely, the likelihood that they will find a solution, if one exists, approaches certainty as computation time increases.

Recent variations of these methods, such as \algname{PRM*} and \algname{RRT*}~\cite{KF11}, improve upon this guarantee by exhibiting \emph{asymptotic optimality}---namely that the quality of the solution obtained, given some cost function, approaches the global optimum as computation increases.
Roughly speaking, this is achieved by increasing the (potential) edge set of roadmap vertices considered as its size increases~\cite{KF11, SSH18}.
One such algorithm is the Rapidly-exploring Random Graphs ({\rrg}s)~\cite{KF11} which will be used in our work.
\rrg
combines the exploration strategy of \rrt with an updated connection strategy that allows for cycles in the roadmap.
It requires solving the two-point boundary value problem~\cite{LaValle2006_Book}, which is only available for some robotic systems (including ours).

Guaranteeing asymptotic optimally can come with heavy computational cost.
This inspired work on planners that trade asymptotic optimality guarantees with \emph{asymptotic near optimality} (e.g.,~\cite{LLB16, MB11-ISRR,SH16}).
Asymptotic near optimality states that given an approximation factor $\eps \geq 0$, the solution obtained converges to within a factor of $(1+\eps)$ of the optimal solution with probability one, as the number of samples tends to infinity.
Relaxing optimality to near optimality allows a method to improve the practical convergence rate while maintaining desired theoretic guarantees on the quality of the solution.

\subsection{Inspection planning}

Many inspection-planning algorithms, or coverage planners,
decompose the region containing the POI into multiple sub-regions, and then solve each sub-region separately~\cite{galceran2013RAS}.
These methods have limitations, however, such as 
when occlusions play a significant role in the inspection~\cite{englot2012_icaps}, or when 
kinematic constraints must be considered~\cite{edelkamp2017RAL}.

Other approaches simultaneously consider all POI.
One approach decouples the problem into the coverage sampling problem (CSP) and the multi-goal planning problem (MPP), and solves each independently~\cite{bircher2015structural,DK00, edelkamp2017RAL, englot2012_icaps, EH11}.
In CSP, a minimal set of viewpoints that provide full inspection coverage is computed.
In MPP, a shortest tour that connects all the viewpoints is computed.
These corresponds to solving the art gallery problem and  the traveling salesman problem, respectively.
Several of these variants have been shown to be probabilistically complete~\cite{englot2012_icaps}, but none provide guarantees on the quality of the final solution.

The set of viewpoints and the inspection plan itself can also be generated simultaneously.
Papadopoulos et al. propose the Random Inspection Tree Algorithm (\rita)~\cite{papadopoulos2013asymptotically}. \rita takes into account differential constraints of the robot and computes both target points for inspection and the trajectory to visit the targets simultaneously.
Bircher et al. propose Rapidly-exploring Random Tree Of Trees (\rrtot) which constructs a meta–tree structure consisting of multiple \algname{RRT*} trees~\cite{BASOBS17}.
Both methods, which were  shown to be asymptotically optimal, iteratively generate a tree, in which the inspection plan is enforced to be a plan from the root to a leaf node.
However, each inspection plan does not consider configurations from other branches in the tree which may cause long planning times.
This motivates our \rrg-based approach.

\subsection{Path planning on graphs}
Planning a minimal-cost path on a graph is a well studied problem.
When the cost function has an optimal substructure (namely, when subpaths of an optimal path are also  optimal), efficient algorithms such as Dijkstra~\cite{D59}, \astar~\cite{HNR68} and the many variants there-of can be used.
However, in certain settings, including ours, this is not the case. 
For example Tsaggouris and Zaroliagis~\cite{TZ04} consider the case where every edge has two attributes (e.g., cost and resource),
and the cost function incorporates the attributes in a  non-linear fashion.

Inspection planning also bears resemblance to multi-objective path planning. Here, we are given a set of cost functions and are required to find the set of Pareto-optimal paths~\cite{pardalos2008pareto}.
Unfortunately, this set may be exponential in the problem size~\cite{RG00}.
However it is possible to compute a fully polynomial-time approximation scheme (FPTAS) for many cases~\cite{TZ06}.
For additional results on path-planning with multiple-objectives or when the cost function does not have an optimal substructure,
see e.g.,~\cite{CN13,RP11} and references within.

\section{Problem Definition}
\label{sec:pdef}

In this section we formally define the inspection planning problem.
The robot operates in a workspace~$\W \subset \mathbb{R}^3$ amidst a set of obstacles~$\W_{\rm obs} \subset \W$. The robot's configuration $\mathbf{q}$ is a $d$-dimensional vector that uniquely defines the shape of the robot (including, for example, rotation angles and translational extension of all joints). 
The set of all such configurations is the configuration space~$\X$.
The geometry of the robot is a configuration-dependent shape~${\rm Shape}(\mathbf{q}) \subset \mathcal{W}$ and we say that $\mathbf{q} \in \X$ is \emph{collision free} if~${\rm Shape}(\mathbf{q}) \cap \W_{\rm obs} = \emptyset$.
In this work we define a motion plan for the robot as a path~$P$ in~$\X$, which is represented as a sequence of~$n$ configurations $\{\mathbf{q}_0, \dots, \mathbf{q}_{n-1}\}$ (vertices) connected by straight-line segments (edges) in~$\X$.
And we say that~$P$ is collision free if all configurations along $P$ (vertices and edges) are collision free.
We assume that we have a distance function $\ell:\X \times \X \rightarrow \mathbb{R}$ and denote the length of a path~$P$ as the sum of the distances between consecutive vertices, i.e., $\ell(P):=\sum_i \ell(\mathbf{q}_i, \mathbf{q}_{i-1})$.

We assume that the robot is equipped with a sensor~$\S$ and we are given a set of $k$ points of interest (POI)~$\I= \{i_1, \ldots, i_k\}$ in~$\W$.
We model the sensor as a mapping~$\S: \X \rightarrow 2^\I$, where~$2^\I$ is the power set of~$\I$ and $S$ denotes the subset of $\I$ that can be inspected from each configuration.
By a slight abuse of notation, given a path~$P$ we set 
$\S(P) :=   \bigcup_{i=0}^{n-1}\S(\mathbf{q}_i)$ and note that in our model, we only inspect $\R$ along the vertices of a path.

\vspace{2mm}
\begin{defin}
	A point of interest $i \in \I$ is said to be \emph{covered} by 
	a configuration $\mathbf{q} \in \X$ or by a path~$P$
	if $i \in \S(\mathbf{q})$ or if $i \in \S(P)$, respectively.
	In such a setting, we say that $\mathbf{q}$ (or~$P$) covers the point of interest~$i$.
\end{defin}
\vspace{2mm}

Given a start configuration~$\mathbf{q}_{\rm s} \in \X$, POI~$\I$, and a sensor model $\S$, the inspection planning problem calls for computing a collision-free path~$P$ starting at~$\mathbf{q}_{\rm s}$  which maximizes~$|\S(P)|$ while minimizing~$\ell(P)$.
Note that this is \emph{not} a bicriteria optimization problem---our primary optimization function is maximizing the coverage of our path.
Out of all such paths we are interested in the shortest one.

\section{Method Overview}
In this section we provide an overview of \iris---our algorithmic framework for computing asymptotically-optimal inspection plans.
A key algorithmic tool in our approach is to cast the continuous inspection planning problem (Sec.~\ref{sec:pdef}) 
to a discrete version of the problem where we only consider a finite number of configurations from which we inspect the POI, and a discrete 
set of feasible movements between those configurations. 
Thus, we start in Sec.~\ref{subsec:graph_inspection} by formally defining the \emph{graph inspection problem} and then continue in Sec.~\ref{subsec:overview} to provide an overview of how \iris builds and uses such graphs. We then describe the method in detail in Sec.~\ref{sec:method}, and in Sec.~\ref{sec:theory} show that \iris's solution converges to the length and coverage of an optimal inspection path. 

\subsection{Graph inspection problem}
\label{subsec:graph_inspection}

Similar to the (continuous) inspection problem, a \emph{graph inspection problem} is a tuple $\left(\G, \R, \S, \ell, v_{\rm s} \right)$ where 
$\G = (\V, \E)$ is a motion-planning roadmap 
(namely, a graph embedded in~$\X$, in which every vertex~$v \in\V$ is a configuration and every edge~$(u,v) \in\E$ denotes the transition from configuration~$u$ to~$v$), 
$\R$ and $\S$ are defined as in Sec.~\ref{sec:pdef},
$\ell: \E \rightarrow \mathbb{R}$ denotes the length of each edge in the roadmap, and
$v_{\rm s}$ is the start vertex 
(corresponding to the start configuration $\mathbf{q}_{\rm s}$).
A path~$P$ on $\G$ is represented by a sequence of vertices~$v_i \in \V$ such that~$P = \{v_0, \dots, v_{n-1} \}$, $v_0 = v_s$ and $(v_{i}, v_{i+1}) \in \E$.
It is important to note that there can be loops in a path, so it is possible that~$v_m = v_k$ for $m \neq k$.
The \emph{length} and \emph{coverage} of~$P$ are defined as the total length of~$P$'s edges and the set of all points inspected when traversing~$P$, respectively.
Namely, 
$\ell(P) := \sum_{i = 0}^{n-2}\ell\left(v_i, v_{i+1}\right)$
and
$\S(P):= \bigcup_{v \in P}\S(v)$. 
The \emph{optimal} graph inspection problem calls for a path~$P^*$ that starts at $v_{\rm s}$ and maximizes the number of points inspected.
Out of all such paths, $P^*$  has the  minimal length.
Finally, a path is said to be \emph{near-optimal} for some $\eps \geq 0$ and $p \in (0,1]$ if
$|\S(P)| / |\S(P^*)| \geq p$
and
$\ell(P) \leq (1 + \eps) \cdot \ell(P^*)$.

\subsection{Overview of \iris}
\label{subsec:overview}

Our algorithmic framework, depicted in Fig.~\ref{fig:method}, interleaves sampling-based motion planning and graph search.
Specifically, we incrementally construct an \rrt~$\T$ rooted at~$\mathbf{q}_{\rm s}$ which implicitly defines a corresponding \rrg~$\G$. 
All edges in~$\T$ are checked for collision with the environment during its construction (so the roadmap is guaranteed to be connected) while all the other edges of~$\G$ are not explicitly checked for collision.
Lazy edge evaluation, common in motion-planning algorithms~\cite{BK00,H15,DS16,SH15a}, allows us to defer collision detection until absolutely necessary and reduce computational effort. 
This is critical in our domain of interest where computing $\rm{Shape}()$ typically dominates algorithms' running times~\cite{NKSAS18}.

The roadmap $\G = (\V, \E)$ induces the subset of the POI that can be inspected, denoted as $\I_\G := \bigcup_{v \in \V} \S(v)$.
Given two approximation parameters~$\eps \geq 0$ and~$p \in (0, 1]$, 
we compute a near-optimal inspection path for the graph-inspection problem~$\left(\G, \R_\G, \S, \ell, v_{\rm s}\right)$ by casting the problem as a graph-search problem on a different graph $\G_{\S}$ (to be defined shortly). 

As we add vertices and edges to~$\T$ incrementally, the roadmap~$\G$ is incrementally densified. 
In addition, we tighten approximations by decreasing~$\eps$ and increasing~$p$ between iterations. As we will see (Sec.~\ref{sec:theory}), this will ensure that \iris is asymptotically optimal.

\section{Method}
\label{sec:method}
In this section we detail the different components of \iris.
Sec.~\ref{subsec:graph construction} and~\ref{subsec:graph search}  describe how we construct a roadmap and then search it, respectively.
After describing in Sec~\ref{subsec:params} how we modify the approximation parameters used by \iris,
we conclude in Sec.~\ref{subsec:implementation}  with implementation details.

\subsection{Roadmap construction}
\label{subsec:graph construction}
We construct a sequence of graphs embedded in~$\X$.
Specifically, denote the \rrt constructed at the $i$'th iteration as~$\T_i$ defined over the set of vertices $\V_i$.
We start with an empty tree rooted at~$\mathbf{q}_{\rm s}$ and at the $i$'th iteration sample a random configuration, compute it's nearest neighbor in~$\T_i$, and extend that vertex toward the random configuration. 
If that extension is collision free we add it to the tree. If not, we repeat this process (see~\cite{Kuffner2000_ICRA,LaValle2006_Book} for additional details regarding \rrt).

The tree~$\T_i$ implicitly-defines an \rrg~$G_i = (\V_i, \E_i)$  defined over the same set of vertices where every vertex is connected to all other vertices within distance~$r_i$.
Here, we define $r_i$ as in~\cite[Thm.~IV.5]{SJP15} which will allow us to prove that our approach is asymptotically optimal (see Sec.~\ref{sec:theory}).

\subsection{Graph inspection planning}
\label{subsec:graph search}

We use the \rrg described in Sec.~\ref{subsec:graph construction} to define a graph inspection problem, and then compute near-optimal inspection paths over this graph. 
Before describing how we compute near-optimal inspection paths, we first describe how we compute optimal paths given a graph inspection problem.

\subsubsection{Optimal planning}
Given a graph inspection problem~$\left(\G, \R_\G, \S, \ell, v_{\rm s}\right)$, we compute optimal inspection paths by formulating our inspection problem as a graph-search problem on an \emph{inspection graph}~$\G_\S := (\V_\S, \E_\S)$.
Here, vertices are pairs comprised of a vertex~$u \in \V$ in the original graph and subsets of $\R_\G$.
Namely, 
$\V_\S = \V \times 2^{\R_\G}$,
and note that~$\vert \V_\S \vert =  O\left(\vert \V \vert \cdot 2^{|\R_\G|} \right)$.
An edge~$e$ between vertices~$(u, \R_u)$ and $(v, \R_v)$ exists if~$(u,v) \in \E$ and~$\R_u \cup \S(v) = \R_v$.
If it exists, its cost is simply~$\ell(u,v)$.

\begin{figure}[tb]
  \centering
 	\includegraphics[height=4.cm]{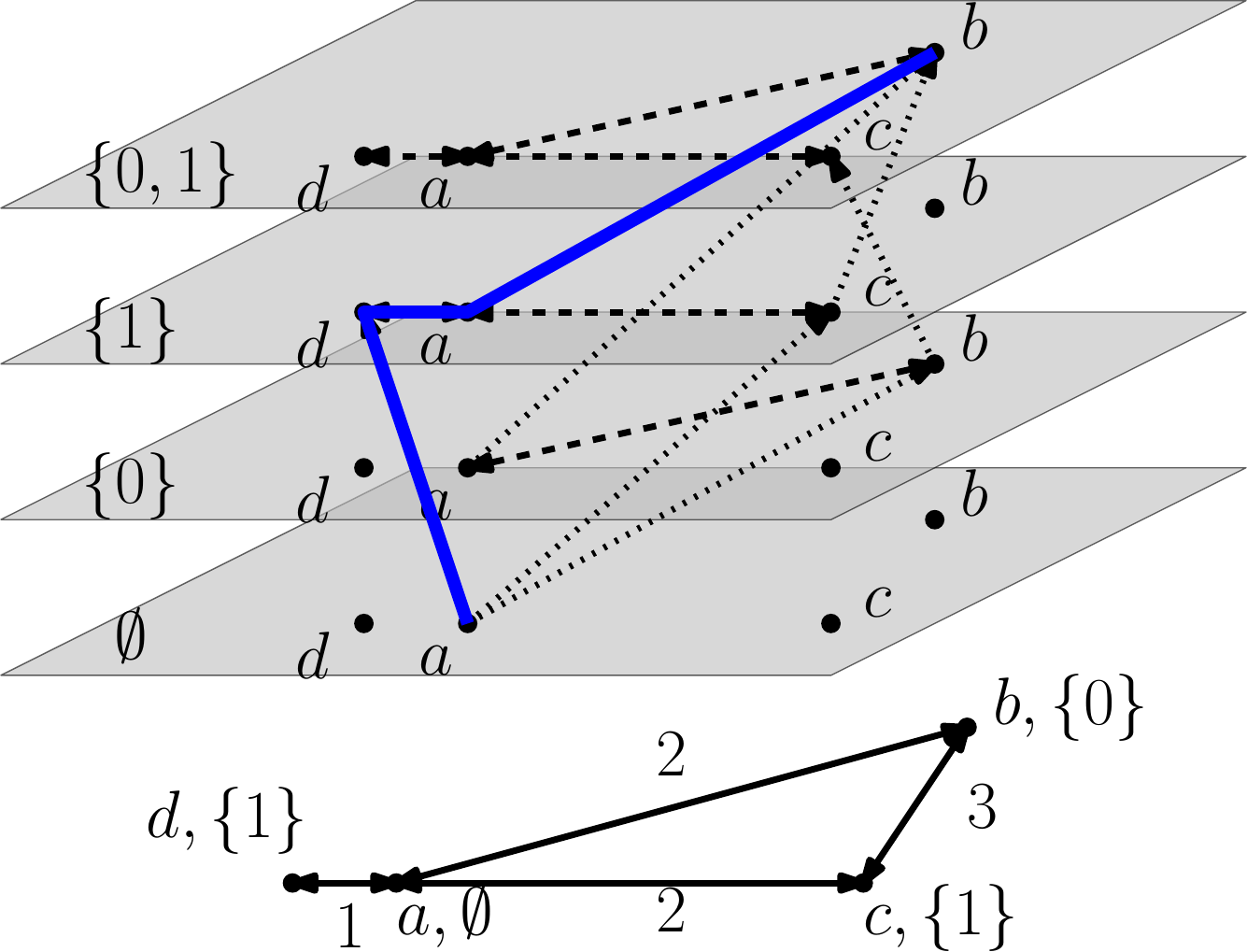}
  \caption{	Computing optimal inspection paths on graphs by casting a graph-inspection problem (bottom) to a graph-search problem (top).
  Grey layers corresponds to the set of all vertices in $\V_\S$ that share the same 
  set of points inspected.
  Edges connecting vertices in the same (different) layer are depicted in dashed (dotted) lines, respectively.
  The start is $(a, \emptyset)$ and the goal set $\V_{\rm goal}$ contains all vertices in the top layer.
  Notice that the optimal path (blue)  visits vertex $a$ twice.}
 \label{fig:exhaustive}
\end{figure}

Any (possibly non-simple) path~$P_G$ in the original graph~$G$ from~$v_{\rm s}$ to~$u$ 
can be represented by a corresponding path~$P_{\G_\S}$ in the inspection graph~$\G_\S$, from~$(v_{\rm s}, \S(v_{\rm s})) \in \V_\S$ to~$(u, \S(P_G)) \in \V_\S$,
and~$\ell(P_G) = \ell(P_{\G_\S})$.
Thus, our algorithm will run an \astar-like search from $(v_{\rm s}, \S(v_{\rm s})) \in \V_\S$ to any vertex in the goal set 
$\V_{\rm goal} = \{ (v, \R_\G) \vert v \in  \V \}$.
An optimal inspection path is the shortest path between $(v_{\rm s}, \S(v_{\rm s}))$ and any vertex in $\V_{\rm goal}$.
For a visualization, see Fig.~\ref{fig:exhaustive}.
Note that here we assume
the graph $\G$ is connected and that the set of points to be inspected is $\R_\G$.
This implies that an optimal inspection path always exists. 

We can speed up this \naive algorithm using the notion of \emph{dominance}, which is used in many shortest-path algorithms~(see, e.g.,~\cite{SHS17}).
In our context, given two paths~$P,P'$ in our original roadmap~$\G$ that start and end at the same vertices, we say that~$P$ \emph{dominates}~$P'$ if~$\ell(P) \leq \ell(P')$ and~$\S(P) \supseteq \S(P')$.
Clearly,~$P$ is always preferred over~$P'$. Thus, when searching~$\G_\S$, if we compute a shortest path to some node~$(u, \R_u)$ of length~$\ell_u$, we do not need to consider any path of length larger than~$\ell_u$ from all vertices~$(u, \R_u')$ such that~$\R_u' \subseteq \R_u$.
For pseudo-code describing a general \astar-like search algorithm to optimally solve the graph-inspection problem, see Alg.~\ref{alg:near-optimal} \emph{without} lines~\ref{alg:line:subsumed1}-\ref{alg:line:dominate2b_}.

While path domination may prune away paths in the open list of the \astar-like search, this algorithm is hardly practical due to the exponential size of the search space (recall that~$|\V_\S| = O(|\V_\S| \cdot 2^{|\R_\G|})$).
In the next sections, we show how to prune away large portions of the search space by extending the notion of dominance to \emph{approximate dominance}.

\subsubsection{Near-optimal planning}
\begin{figure*}[tbh]
\begin{subfigure}{.245\linewidth}
    \centering
    {\includegraphics[width=0.975\linewidth]{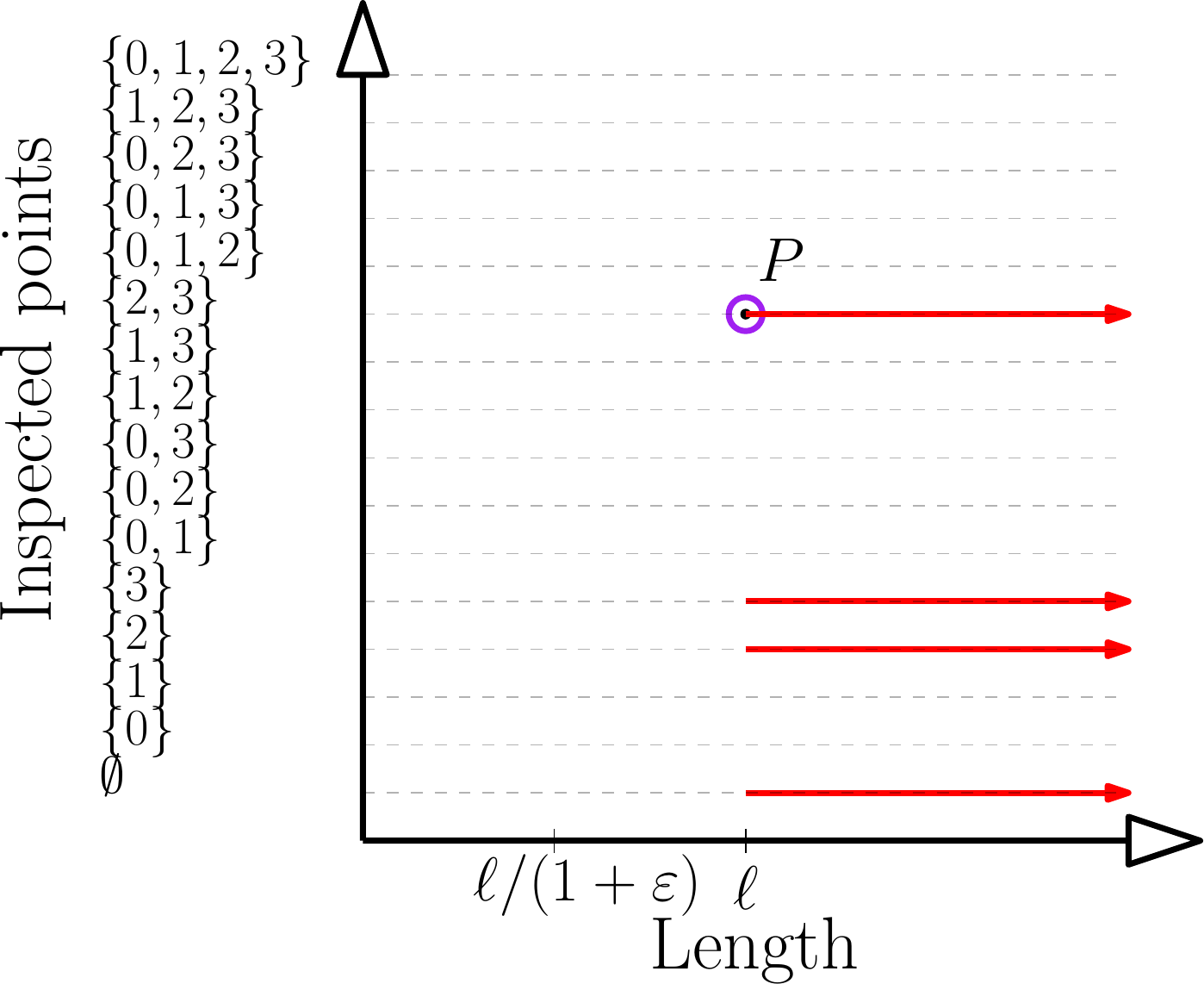}}
 \caption{Dominance}
 \label{fig:s-dominance}
\end{subfigure}
\begin{subfigure}{.245\linewidth}
    \centering
    {\includegraphics[width=0.975\linewidth]{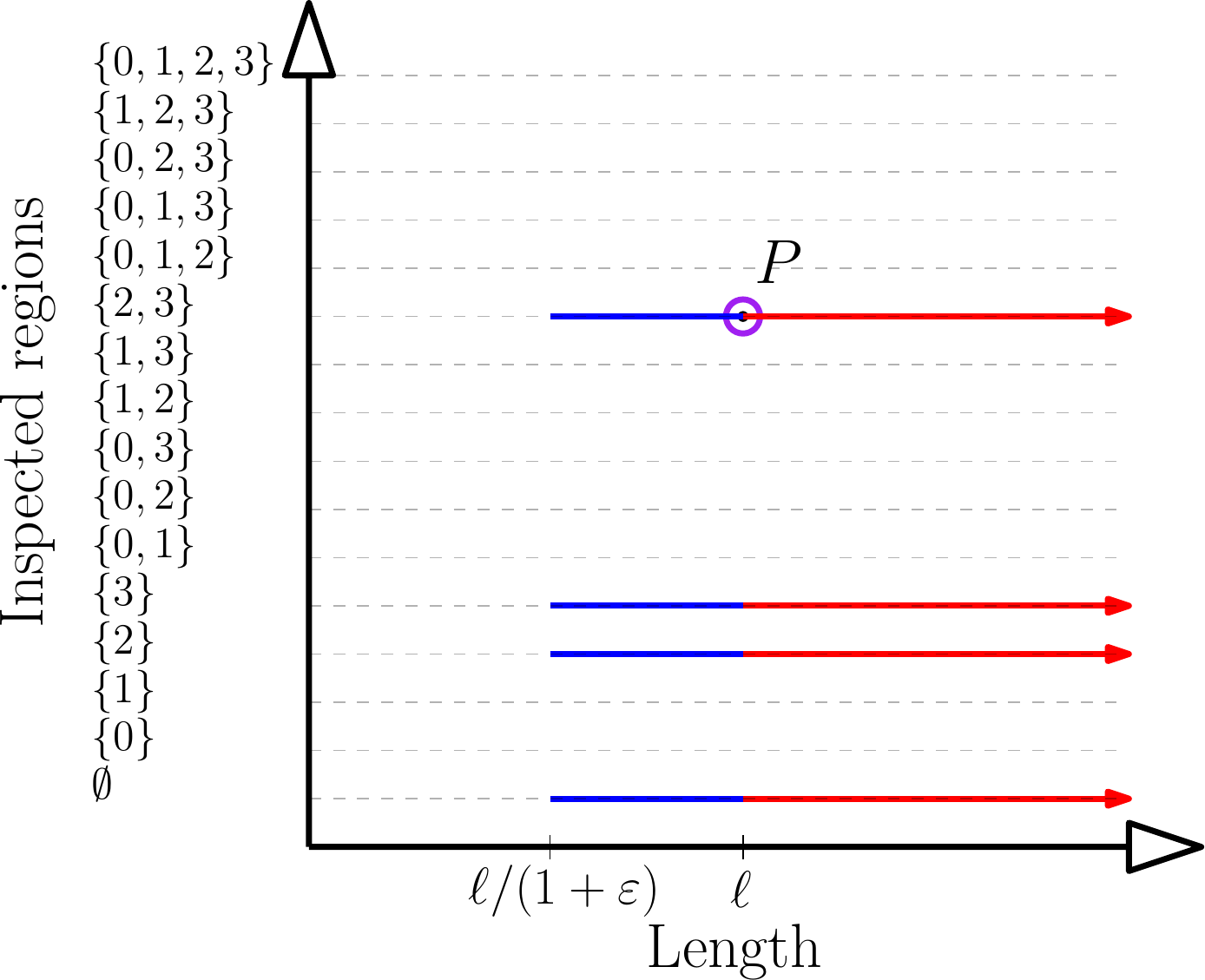}}
 \caption{$\eps$-dominance}
 \label{fig:e-dominance}
\end{subfigure}
\begin{subfigure}{.245\linewidth}
    \centering
    {\includegraphics[width=0.975\linewidth]{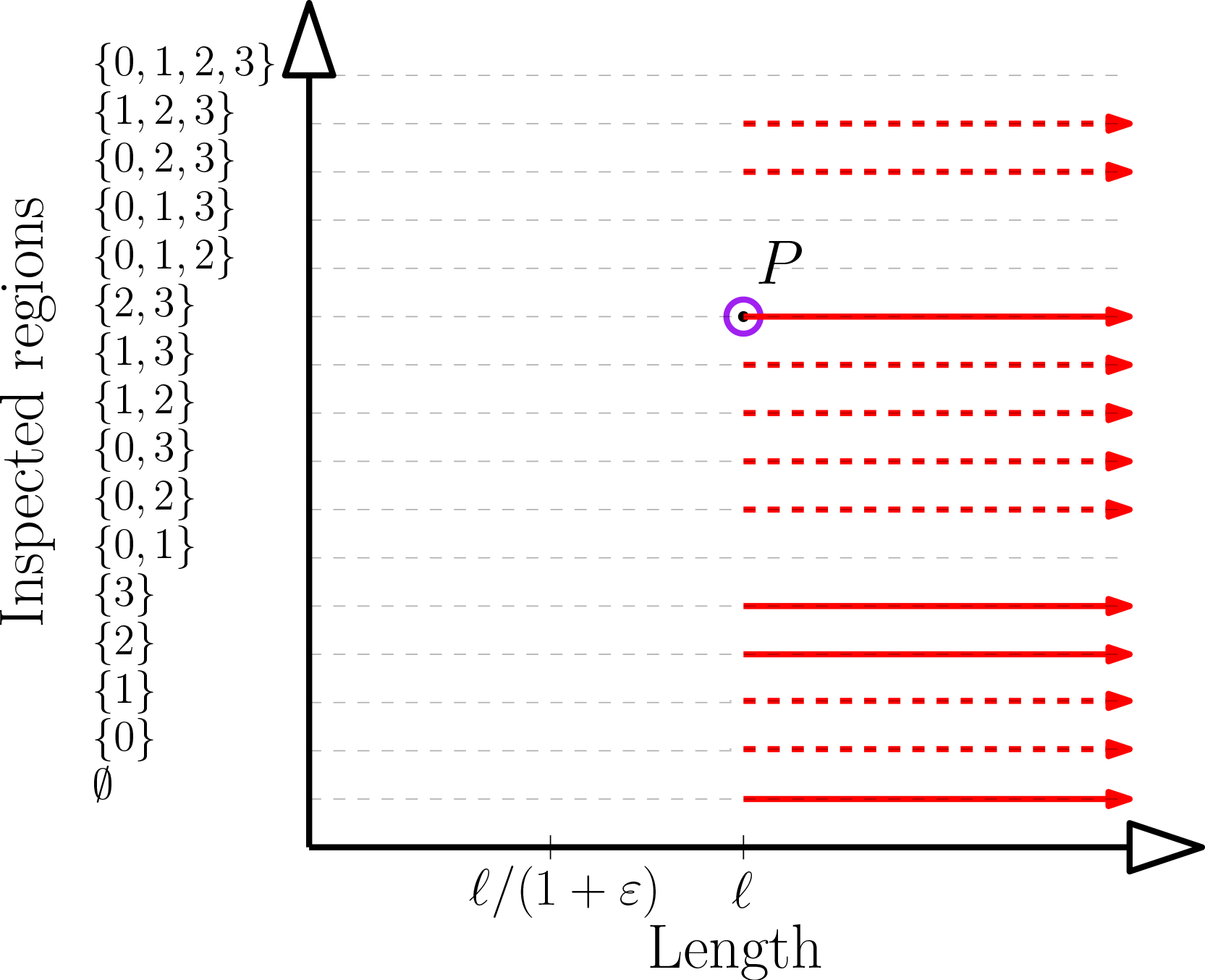}}
 \caption{$p$-dominance}
 \label{fig:p-dominance}
\end{subfigure}
\begin{subfigure}{.245\linewidth}
    \centering
    {\includegraphics[width=0.975\linewidth]{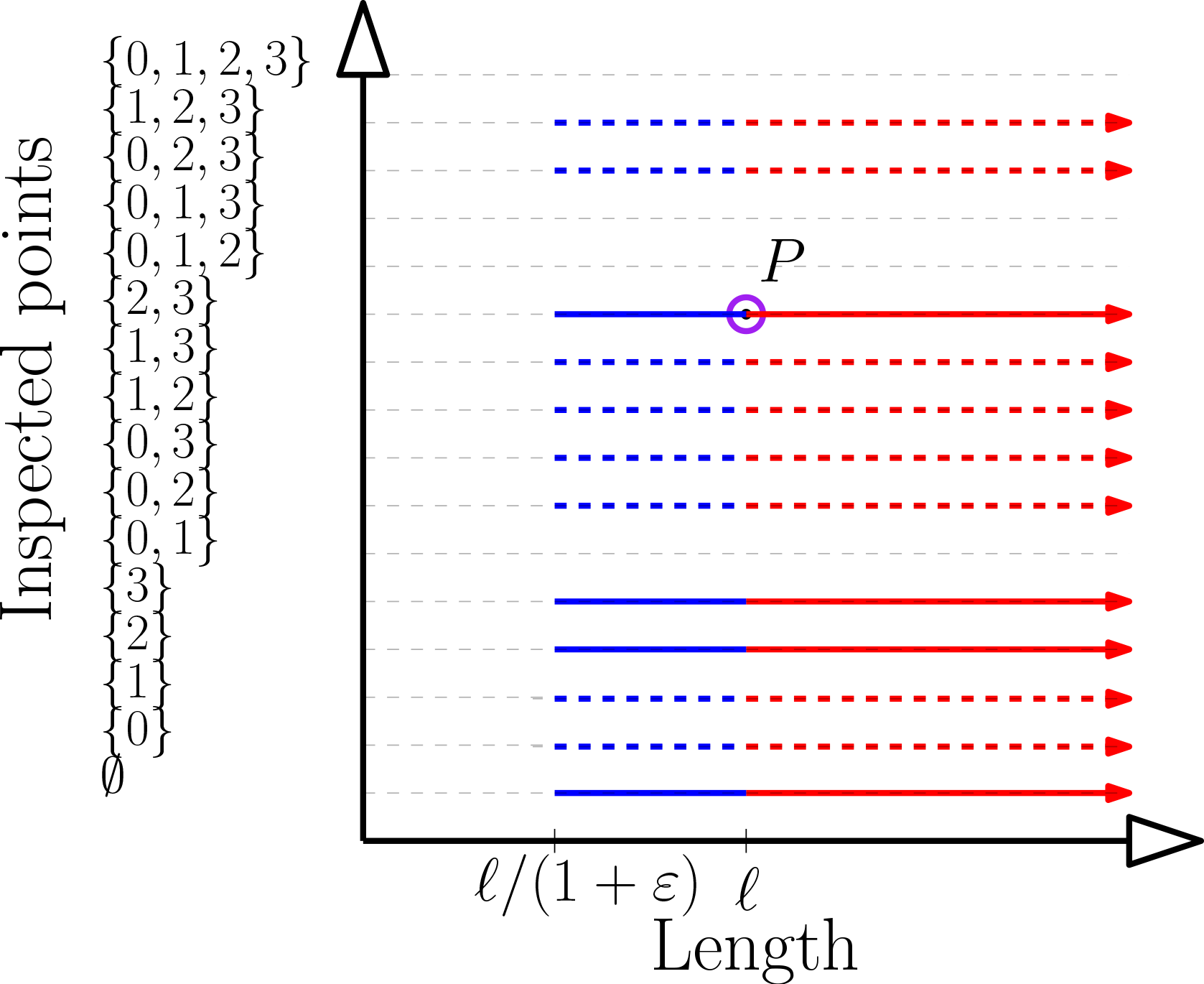}}
 \caption{$(\varepsilon,p)$-dominance}
 \label{fig:ep-dominance}
\end{subfigure}

    \caption{
    Visualization of the notion of dominating paths by considering a path~$P$ from~$v_{\rm s}$ to some vertex~$u$ as a two-dimensional point~$(\ell(P),\S(P))$.
	Here $\R_\G = \{0,1,2,3\}$ and $P$ is depicted using the purple circle with $\ell(P) = \ell$ and $\S(P) = \{ 2,3\}$.
	All paths from $v_{\rm s}$ to $u$ that
 \protect (\subref{fig:s-dominance})~are dominated by $P$ (solid red),
 \protect (\subref{fig:e-dominance})~are $\varepsilon$-dominated by~$P$ (solid blue),
 \protect (\subref{fig:p-dominance})~are $p$-dominated by $P$ for $p=60\%$ (dashed red),
 \protect (\subref{fig:ep-dominance})~are $\eps,p$-dominated by $P$ for~$\eps>0$ and $p = 60\%$ (dashed blue).
 }
\label{fig:domination}
\end{figure*}

Let~$P,P'$ be two paths in~$\G$ that start and end at the same vertices and let~$\varepsilon \geq 0$ and~$p \in (0,1]$ be some approximation parameters. 

\vspace{2mm}
\begin{defin}
We say that path $P$ \emph{$\varepsilon,p$-dominates} path~$P'$ if~$\ell(P) \leq  (1+\varepsilon) \cdot \ell(P')$
and~$|\S(P)|  \geq p \cdot |\S(P) \cup \S(P')|$. 
\end{defin}
\vspace{2mm}

Note that it is possible that
both
$P$ $\varepsilon,p$-dominates  $P'$ and~$P'$ $\varepsilon,p$-dominates $P$.
For a visualization of the notions of dominance and the approximate dominance, see Fig.~\ref{fig:domination}.

Intuitively, approximate dominance allows to dramatically prune the search space by only considering paths that can significantly improve the quality (either in terms of length or the set of points inspected) of a given path.
When pruning away (strongly-) dominated paths, it is clear that they cannot be useful in the future.
However, if we prune away approximate-dominated paths, we need to  efficiently account for all paths that were pruned away in order to bound the quality of the solution obtained.
We do this using the notion of \emph{potentially-achievable paths} or \paps.

\vspace{2mm}
\begin{defin}
	A potentially-achievable path (\pap)~$\tilde{P}$ to some vertex~$u \in \V$ is a pair
	$(\tilde{\ell}, \tilde{\R})$ such that $\tilde{\ell} \geq 0$ and $\S(u) \subseteq \tilde{\R} \subseteq \R_\G$.
	By a slight abuse of notation, we extend the definitions of $\ell(\cdot)$ and $\S(\cdot)$ such that 
	$\ell(\tilde{P}) = \tilde{\ell}$ and 
	$\S(\tilde{P}) = \tilde{\R}$.
\end{defin}
\vspace{2mm}

It may seem that a \pap is simply a path but note (as the name \pap suggests) that we do not require that there exists any path~$P$ from $v_{\rm{s}}$ to~$u$ such that 
$\ell(P) = \ell(\tilde{P})$ and 
$\S(P) = \S(\tilde{P})$.
It merely states that such a path \emph{may} exist.

We now use \paps to define the notion of a \emph{path pair}:

\vspace{2mm}
\begin{defin}
    Let $P$ and~$\tilde{P}$ be a path and a \pap from $v_{\rm{s}}$ to some $v \in \V$ such that
	$\ell(\tilde{P}) \leq \ell(P)$ 
	and
	$\S(\tilde{P}) \supseteq \S(P)$.
	Their path pair is \pp$:=(P, \tilde{P})$  and we call $P$ and~$\tilde{P}$ the achievable and potentially-achievable paths of \pp, respectively.
\end{defin}
\vspace{2mm}

\begin{figure}[t!]
\begin{subfigure}{.495\linewidth}
  \centering
  {\includegraphics[width=0.975\linewidth]{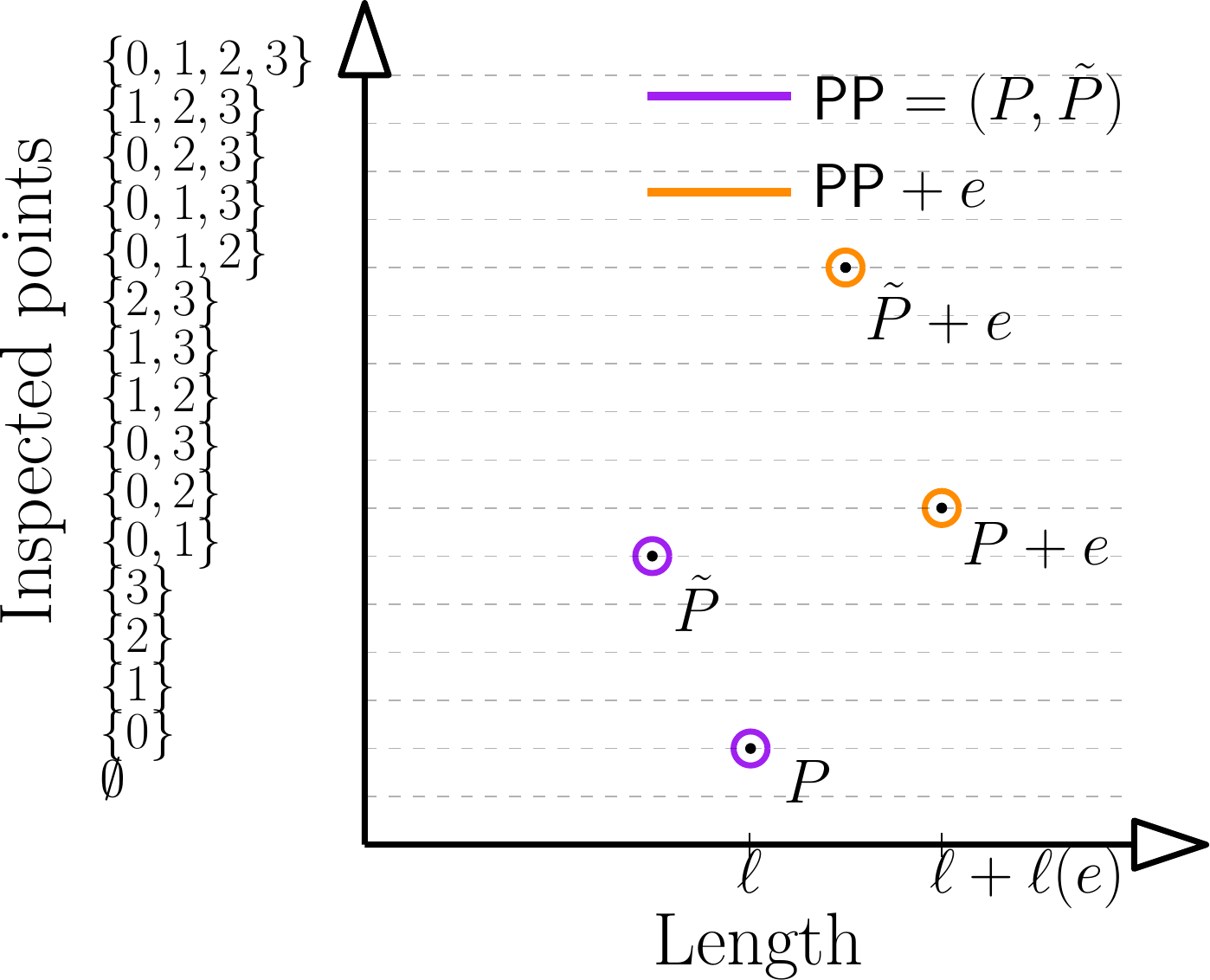}}
  \caption{Extending a path pair}
  \label{fig:extend}
\end{subfigure}
\begin{subfigure}{.495\linewidth}
  \centering
  {\includegraphics[width=0.975\linewidth]{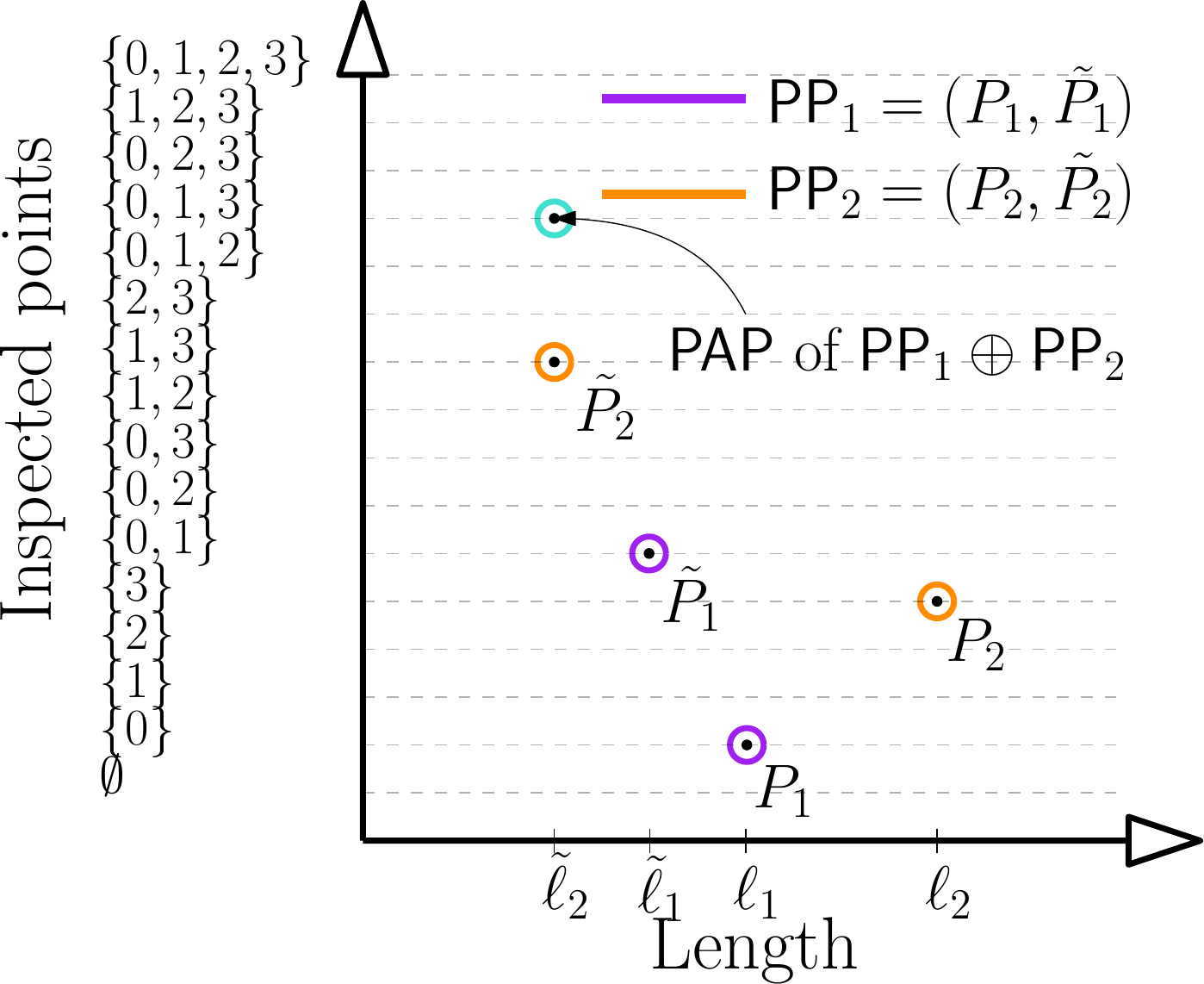}}
  \caption{Subsuming a path pair}
  \label{fig:subsume}
\end{subfigure}
  \caption{Depiction of operations on path pairs. 
  \protect (\subref{fig:extend})~\pp extended by an edge $e = (u,v)$ with $\S(v) = \{ 2\}$.
  \protect (\subref{fig:subsume})~$\pp_1$ subsuming $\pp_2$. Note that $P_1$ is the achievable path of $\pp_1 \oplus \pp_2$ thus only the  potentially-achievable path is explicitly marked.
  }
 \label{fig:pp-operations}
\end{figure}

Let us define operations on \paps and on \pps, visualized in Fig.~\ref{fig:pp-operations}.
The first operation we consider is \emph{extending} a \pap~$\tilde{P_u}$ by an edge~$e = (u,v)$, denoted as $\tilde{P_u} +e$.
This can be thought of as appending~$e$ to $\tilde{P_u}$, had it existed and thus accounting for the length $\ell(e)$ and additional coverage~$\S(v)$.
Formally, extending $\tilde{P_u} +e$ yields a \pap $\tilde{P_v}$ such that
$\ell(\tilde{P_u}) = \ell(\tilde{P_v}) + \ell(e)$
and
$\S(\tilde{P_u}) = \S(\tilde{P_v}) \cup \S(u)$.
Extending the path pair~$\pp_u =(P_u, \tilde{P_u})$ by the edge~$e = (u,v)$ (denoted as $\pp_u + e$) simply extends both~$P_u$ and $\tilde{P_u}$ by~$e$.
This yields the path pair
$\pp_v =(P_v, \tilde{P_v})$ where
$\ell({P_v}) = \ell({P_u}) + \ell(e)$,
$\S({P_v}) = \S({P_u}) \cup \S(v)$
and
$\tilde{P_v} = \tilde{P_u} + e$.

The second operation is \emph{subsuming} a path pair by another one which can be thought of as constructing a PAP that dominates the PAPs of both path pairs.
Formally,
Let 
$\pp_1 = (P_1, \tilde{P_1})$ and 
$\pp_2 = (P_2, \tilde{P_2})$ 
be two path pairs such that both connect the start vertex~$v_{\rm s}$ to some vertex~$u \in \V$. 
The path pair defined by~$\pp_1$ subsuming~$\pp_2$ is
$$
\pp_1 \oplus \pp_2
:=
(
	P_1,
	(
	\min\{\ell(\tilde{P_1}), \ell(\tilde{P_2})\},
	\S(\tilde{P_1}) \cup \S(\tilde{P_2})
	)
).
$$

We now define the notion of bounding a path pair which will be crucial for ensuring near-optimal solutions:

\vspace{2mm}
\begin{defin}
    A path pair $\pp:=(P, \tilde{P})$ is said to be \emph{$\varepsilon,p$-bounded} for some $\varepsilon \geq 0$
    and
    $p \in (0,1]$
    if 
    $P$ $p,\varepsilon$-dominates~$\tilde{P}$.
\end{defin}
\vspace{2mm}

\begin{figure*}[tbh]
\begin{subfigure}{.245\linewidth}
    \centering
     {\includegraphics[width=0.85\linewidth]{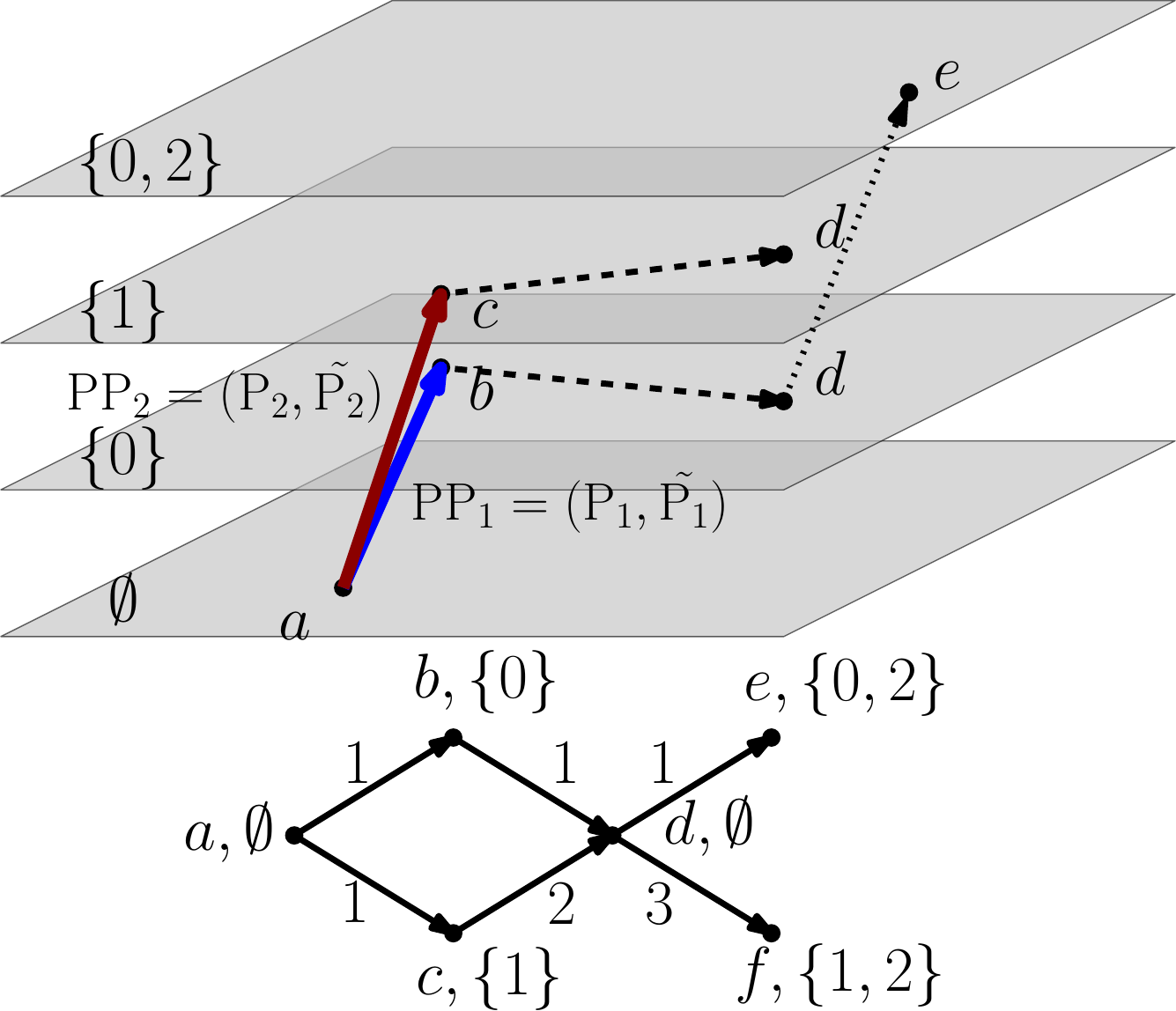}}
  \caption{}
  \label{fig:alg1}
\end{subfigure}
\begin{subfigure}{.245\linewidth}
    \centering
     {\includegraphics[width=0.85\linewidth]{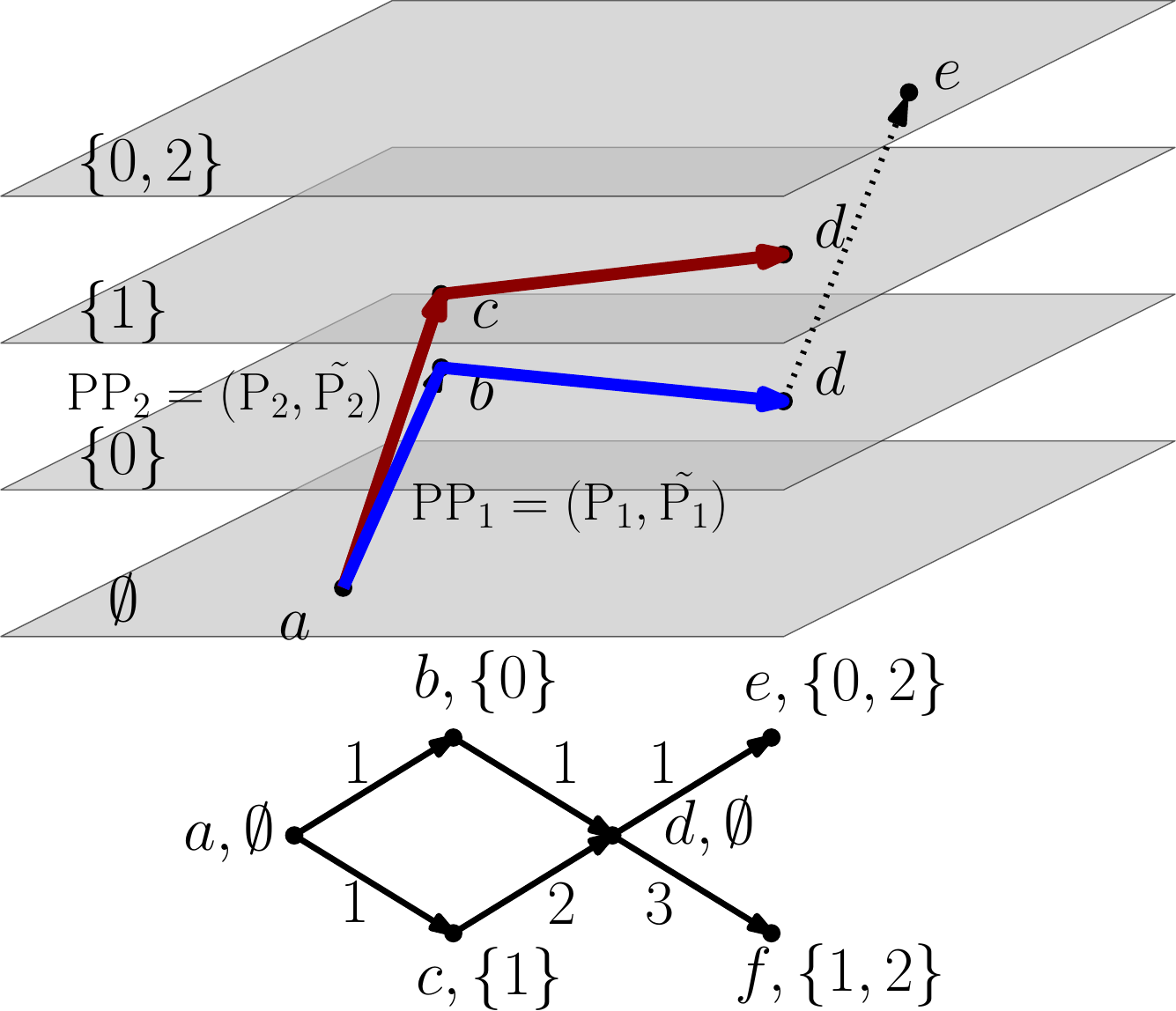}}
  \caption{}
  \label{fig:alg2}
\end{subfigure}
\begin{subfigure}{.245\linewidth}
    \centering
     {\includegraphics[width=0.85\linewidth]{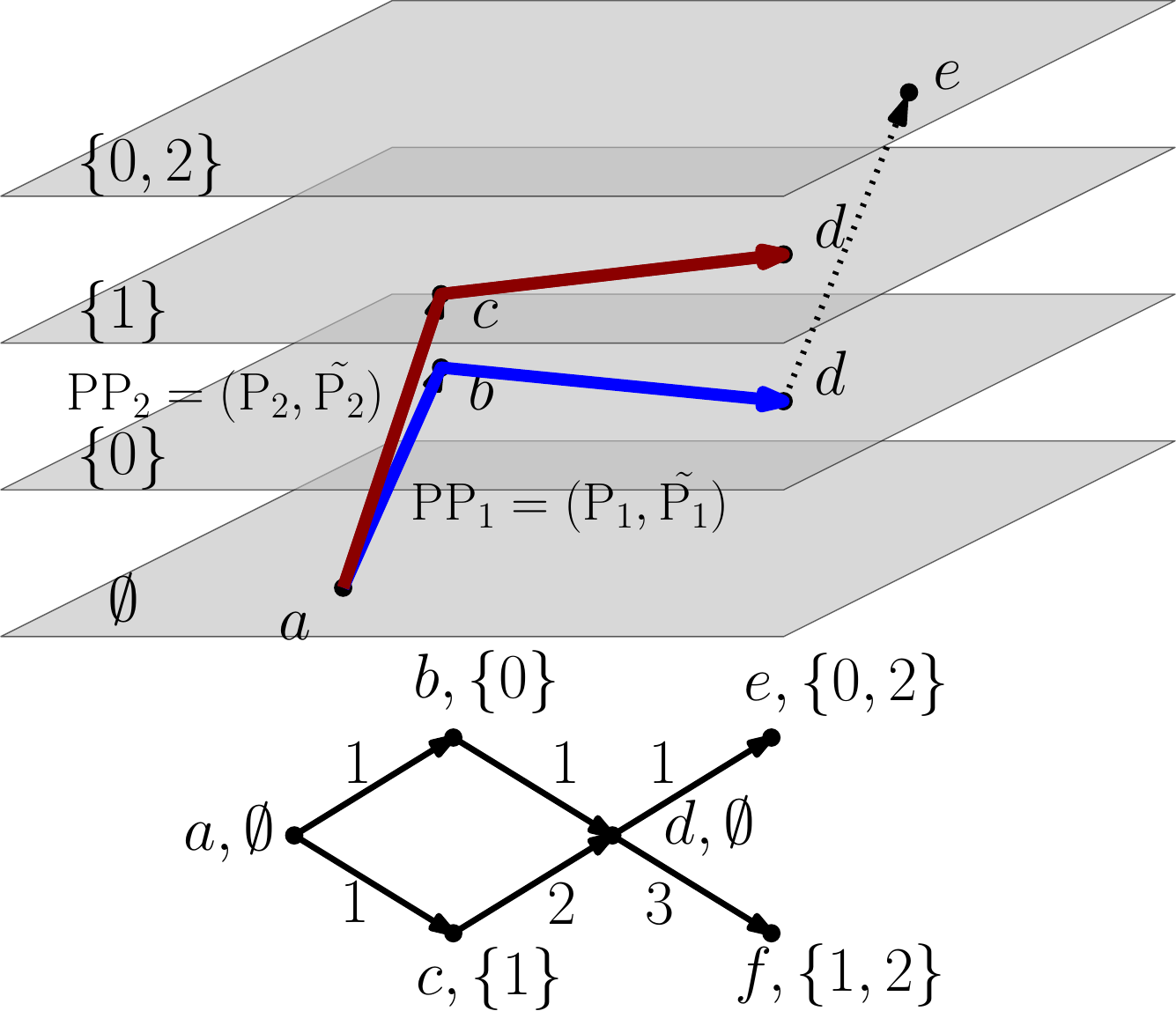}}
  \caption{}
  \label{fig:alg3}
\end{subfigure}
\begin{subfigure}{.245\linewidth}
    \centering
     {\includegraphics[width=0.85\linewidth]{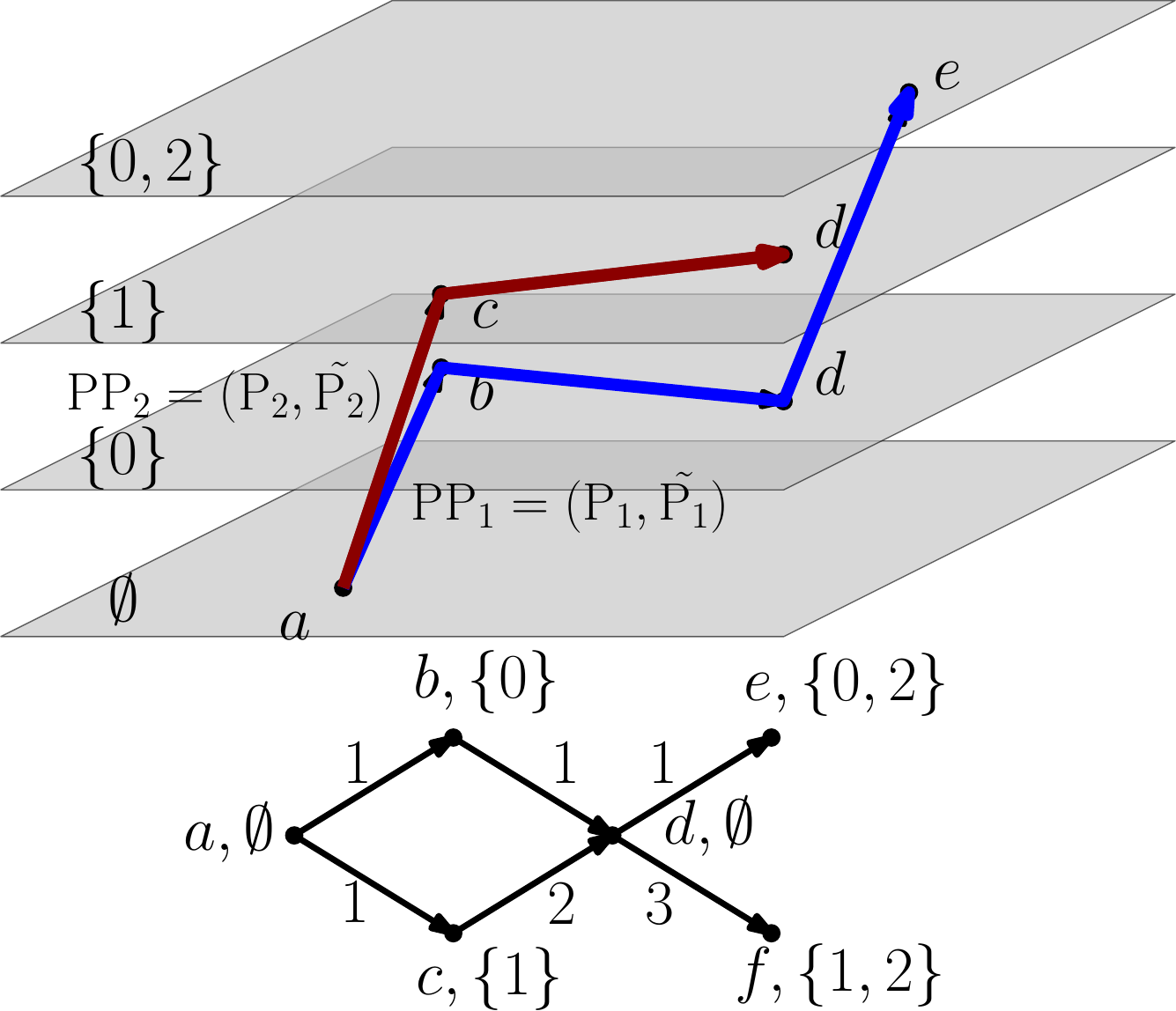}}
  \caption{}
  \label{fig:alg4}
\end{subfigure}
    \caption{
  Visualization of Alg.~\ref{alg:near-optimal} initialized with $\eps = 2/3$ and $p = 1/2$ (only the relevant parts of the inspection graph are depicted).
  The search starts from $(a,\emptyset)$ with the trivial \pap of length zero and 
  no points inspected.
  \protect (\subref{fig:alg1})~Two paths (red and blue) are extended from the start node to $(b,\{0\})$ and $(c,\{1\})$  with path pairs $\pp_1$ and $\pp_2$, respectively (the \paps of each path have the same length and coverage as the paths themselves).
  \protect (\subref{fig:alg2})~Blue path extended to $(d,\{0\})$ with $\ell(P_1) = \ell (\tilde{P_1}) = 2$ and $\S(P_1) = S(\tilde{P_1}) = \{0\}$.
  \protect (\subref{fig:alg3})~Red path extended to $(d,\{1\})$ with $\ell(P_2) = \ell (\tilde{P_2}) = 3$ and $\S(P_2) = S(\tilde{P_2}) = \{1\}$.
  Here, $\pp_1 \oplus \pp_2$ $\eps,p$-dominates $\pp_2$ and the red path is discarded and $\pap_1$ is updated to have length 2 and coverage $\{ 0,1\}$ 
  \protect (\subref{fig:alg4})~Blue path extended to vertex $(e,\{0,2\})$.  Here, $\ell(P_1) = \ell (\tilde{P_1}) = 3$ and $\S(P_1) = \{0,2\},  S(\tilde{P_1}) = \{0,1,2\}$.
  The algorithm terminates with the path $a-b-d-e$ whose length is 3 and has coverage of $\{0,2\}$. Notice that the path $a-c-d-e$ (not computed) is optimal as its length is four and it has complete coverage. 
  The computed path is within the bounds ensured by the approximation factor~$p$ and $\eps$. 
  }
 \label{fig:alg}
\end{figure*}

To compute a near-optimal inspection path (Alg.~\ref{alg:near-optimal} and Fig.~\ref{fig:alg}), we extend each path considered by our search algorithm to be a path pair and use this additional data to prune away approximately-dominated paths.
Similar to \astar, our algorithm uses two priority queues OPEN and CLOSED to track the nodes considered by the search.
It starts by inserting the start vertex $(v_{\rm s}, \S(v_{\rm s}))$ to the OPEN list together with the path pair ${\pp}_s = (P_{\rm s}, P_{\rm s})$ (here~$P_{\rm s}$ is a path containing only start vertex $v_{\rm s}$)  (line~\ref{alg:line:pathset_}).

Our algorithm proceeds in a similar fashion to \astar---we pop the most promising node $n = (u,\R_u, \pp_u)$ from OPEN (line~\ref{alg:line:pop}) and move it to CLOSED (line~\ref{alg:line:closed}).
If the PAP of this node is in the goal set $\V_{{\rm goal}}$ (line~\ref{alg:line:goal_}), we terminate the search and return the achievable path of $\pp_u$ (line~\ref{alg:line:return}).
If not, we consider all  neighboring edges $e$ of $u$ in $\G$ and extend the node~$n$ (line~\ref{alg:line:extendb_}). This is akin to computing $n$'s neighbors in~$\G_\S$.

At this point our algorithm deviates from the standard \astar algorithm.
For each newly-created node $(v,\R_v,\pp_v)$ we check if there exists a node in CLOSED that dominates it. If so, this node is discarded (lines~\ref{alg:line:dominated1}-\ref{alg:line:dominated2}).
If no such node exists in the CLOSED list, we check if there exists a node in OPEN that may subsume it. If so, that node is updated and this node is discarded (line~\ref{alg:line:subsumed1}-\ref{alg:line:subsumed2}).
Finally, we check if this node can subsume nodes that are in OPEN. If so, such nodes are discarded and this node is updated. (line~\ref{alg:line:subsume1}-\ref{alg:line:subsume2}).

It is straightforward to see that 
(i)~the first path pair is $\varepsilon,p$-bounded
and hence by induction 
(ii)~all path pairs in the search are $\varepsilon,p$-bounded.
Furthermore, when the algorithm terminates, it has found a valid solution whose potentially-achievable path is in the goal set.
This yields the following corollary:
\begin{cor}
\label{cor:opt}
    Alg.~\ref{alg:near-optimal} computes a path $P$ that $\varepsilon,p$-dominates the optimal inspection path $P^*$.
    Namely, that
        $\ell(P) \leq (1 + \varepsilon) \cdot \ell(P^*)$
    and~$|\S(P)|  \geq p \cdot |\S(P) \cup \S(P^*)|$.
\end{cor}

\begin{algorithm}[t!]
    \begin{algorithmic}[1]
        \State {${\rm CLOSED}\leftarrow \emptyset$}
        \State {${\rm OPEN}\leftarrow (v_{\rm s}, \S(v_{\rm s}), {\pp}_{\rm s})$}
            \label{alg:line:pathset_}
        \vspace{2mm}
        \While{${\rm{OPEN}} \neq \emptyset$}
            \State $(u, \R_u, {\pp}_u) \gets$ \rm{OPEN{}.extract\_min}()
            \label{alg:line:pop}
            \State {\rm{CLOSED.insert}}($u, \R_u, \pp_u$)
            \label{alg:line:closed}
             \If{$\S(\tilde{P_u}) \in \V_{\rm goal}$}
                 \Comment{{\footnotesize $\tilde{P_u}$ is the \pap of ${\pp}_u$}}
                \label{alg:line:goal_}
                \State \Return $P_u$ \Comment{{\footnotesize ${P_u}$ is the achievable path  of ${\pp}_u$}}
                \label{alg:line:return}
            \EndIf
        \vspace{2mm}
            \For{$e=(u,v) \in$ {neighbors}($u, \G$)}
\label{alg:neighbor}
                    \State{$(v, \R_v, {\pp}_v)\leftarrow$ \rm{extend$((u, \R_u,{\pp}_u), e)$}} \label{alg:line:extendb_}
\vspace{2mm}
                        \State valid = \textbf{True}
                        \For{$(v, \R_v', {\pp}_v') \in {\rm CLOSED}$}
                        \label{alg:line:dominated1}
                            \If{$P_v'$ dominates $P_v$}
                                \State valid = \textbf{False}
                                \State \textbf{break}
                            \EndIf
                            \label{alg:line:dominated2}
                        \EndFor
        \vspace{2mm}
%
                    % \State valid = \textbf{True}
                    \If{!valid}
                        \State \textbf{continue}
                    \EndIf
        \vspace{2mm}
                        \For{$(v, \R_v', {\pp}_v') \in {\rm OPEN}$}
                        \label{alg:line:subsumed1}
                            \If{
                            ${\pp}_v' \oplus {\pp}_v$ is  $\varepsilon,p$-bounded}
                                \State{$(v, \R_v', {\pp}_v')\leftarrow (v, \R_v', {\pp}_v' \oplus {\pp}_v)$}
                                \State valid = \textbf{False}
                                \State \textbf{break}
                                \label{alg:line:subsumed2}
                            \EndIf
\label{alg:line:dominate1b_}
                    \EndFor
        \vspace{2mm}
                    \If{!valid}
                        \State \textbf{continue}
                    \EndIf
        \vspace{2mm}
                        \For{$(v, \R_v', {\pp}_v') \in {\rm OPEN}$}
                        \label{alg:line:subsume1}
                            \If{
                            ${\pp}_v \oplus {\pp}_v'$ is $\varepsilon,p$-bounded}
                            
                                \State {\rm{OPEN}.remove}$(v, \R_v', {\pp}_v')$ 
                                \State{$(v, \R, {\pp}_v)\leftarrow (v, \R, {\pp}_v \oplus {\pp}_v')$}
                                \label{alg:line:subsume2}
                            \EndIf
\label{alg:line:dominate2b_}
                        \EndFor
        \vspace{2mm}
                        \State {\rm{OPEN}}$\leftarrow (v, \R_v, {\pp}_v)$
                \EndFor
            \EndWhile
            \State \textbf{return}  NULL
    \end{algorithmic}
    \caption{Near-optimal inspection planning\\
            \text{Input:} ($\G_{\S}$, $v_{\rm s}, \V_{\rm goal}$, $\varepsilon,p$)}
    \label{alg:near-optimal}
\end{algorithm}

\subsection{Tightening approximation factors}
\label{subsec:params}
Recall that our algorithm starts with approximation parameters~$p_0$ and~$\eps_0$.
We endow our algorithm with a tightening factor $f \in (0,1]$,
and at the $i$'th iteration we set our approximation parameters as~$p_i = p_{i-1} + f \cdot (1 - p_{i-1})$ and ~$\eps_i = \eps_{i-1} + f \cdot (0 - \eps_{i-1})$.
As we will see (Sec.~\ref{sec:theory}), the tightening allows our method to achieve asymptotic optimality.

\subsection{Implementation details}
\label{subsec:implementation}
\subsubsection{Lazy computation in graph inspection planning}
We run our search algorithm on~$\G$ (Alg.~\ref{alg:near-optimal}) without checking if its edges are collision free or not (recall that only the edges of~$\T$ were explicitly checked for collision).
Once a solution is found, we start checking edges one by one until the entire path was found to be collision free or until one edge is found to be in collision, in which case we remove it from the edge set.
This approach is common to speed up motion-planning algorithms when edges are expensive to evaluate~\cite{DS16,HMPSS18}.

\subsubsection{Node extension in graph inspection planning}
Any optimal inspection path can be decomposed into a sequence of vertices that improve the coverage of the path called \emph{milestones}. 
It is straightforward to see that an optimal inspection path will
(i)~terminate at a milestone and
(ii)~connect a pair of milestones via a shortest path in~$\G$.
Following this observation, instead of extending each path $P$ from a vertex~$u$ by all outgoing edges in $\G$ (Alg.~\ref{alg:near-optimal}, line~\ref{alg:neighbor}), we run a Dijkstra-like search from~$u$ and collect all first-met vertices that can be  milestones.

\subsubsection{Heuristic computation in graph inspection planning}
Recall that \astar orders nodes in the OPEN list according to their computed cost-to-come added to a heuristic estimate of their cost to reach the goal.
The heuristic function that we use for vertex $(u, \R_u)$ is computed as follows: we run a Dijkstra search on~$\G$ from $u$ and consider the vertices $\V_u$ encountered during the search. We terminate when $\left(\bigcup_{v \in \V_u} \S(v)\right) \cup \R_u = \R_G$ and use the maximal distance between $u$ to any node in $\V_u$ as our admissible~\cite{HNR68} heuristic function.

\section{Theoretical Guarantees}
\label{sec:theory}
We provide a proof sketch showing that, asymptotically, the length and coverage of the  path produced by \iris will converge to the length and coverage of an  optimal inspection path.
For ease of exposition, we state our results for the following simplified variant of \iris where we start with an empty roadmap~$G_0$ and some initial approximation factors $\varepsilon_0$ and $p_0$.
At each iteration~$i$ we
(i)~sample a collision-free configuration $q_i$ uniformly at random from~$\X$,
(ii)~add $q_i$ to roadmap $G_i$ and connect it to all samples within radius $r_i$,
and
(iii)~compute a near-optimal inspection path on this roadmap with parameters~$\varepsilon_i$ and $p_i$.
(Namely, instead of implicitly constructing an \rrg, we implicitly construct a \prm.
While not identical, both roadmaps exhibit similar properties which are typically easier to show for \algname{PRM\text{s}}.)

Roughly speaking, the connection radius $r_i$ was chosen to ensure that as $i \rightarrow \infty$ an optimal inspection path may be traced arbitrarily well by the  roadmap~$\{G_i\}_{i=1}^\infty$.
The approximation factors were chosen such that $\varepsilon_i > \varepsilon_{i+1}$, 
$p_i < p_{i+1}$,
$\Lim{i \rightarrow \infty} \varepsilon_i =0$
and
$\Lim{i \rightarrow \infty} p_i =1$.
This will ensure that as $i \rightarrow \infty$ the inspection path found will converge to an optimal inspection path in $G_i$. 

A key result that we rely on is \emph{probabilistic exhaustivity} (see,~\cite[Thm.~IV.5]{SJP15} and~\cite{ISP17,SGSJMP15}).
Roughly speaking, it is the notion that given a sufficiently large set of uniformly sampled configurations, any path can be traced arbitrarily well.

Finally, we assume that for every configuration along an optimal inspection path, there exists a neighborhood of configurations that share the same visibility.
This is critical as we will not be able to exactly trace an optimal inspection path but only to iteratively approximate it.
When this assumption holds we say that our inspection problem is \emph{well behaved}.

The combination of 
(i)~probabilistic exhaustivity,
(ii)~the $\varepsilon,p$-dominance of our graph inspection algorithm (Cor.~\ref{cor:opt}) 
(iii)~that  
$\Lim{i \rightarrow \infty} \varepsilon_i =0$ and 
$\Lim{i \rightarrow \infty} p_i =1$, 
and
(iv)
that our inspection problem is well behaved
ensures that asymptotically, our algorithm will converge to an  optimal inspection path. 

\section{Results}

\begin{figure}
\begin{subfigure}{.38\linewidth}
    \centering
     {\includegraphics[height=3cm]{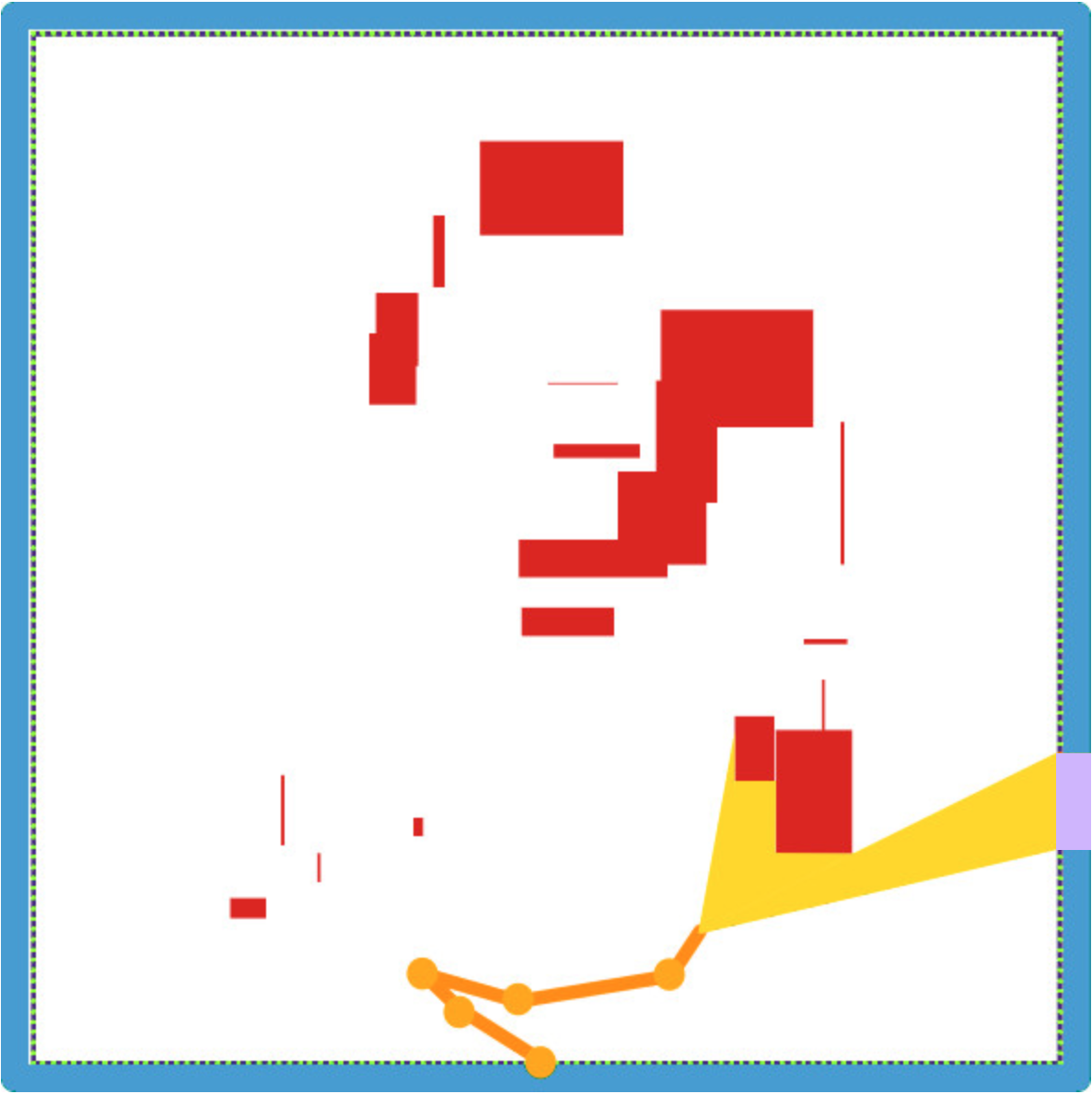}}
  \caption{Planar manipulator}
  \label{fig:planar}
\end{subfigure}
\begin{subfigure}{.610\linewidth}
    \centering
     {\includegraphics[height=3cm]{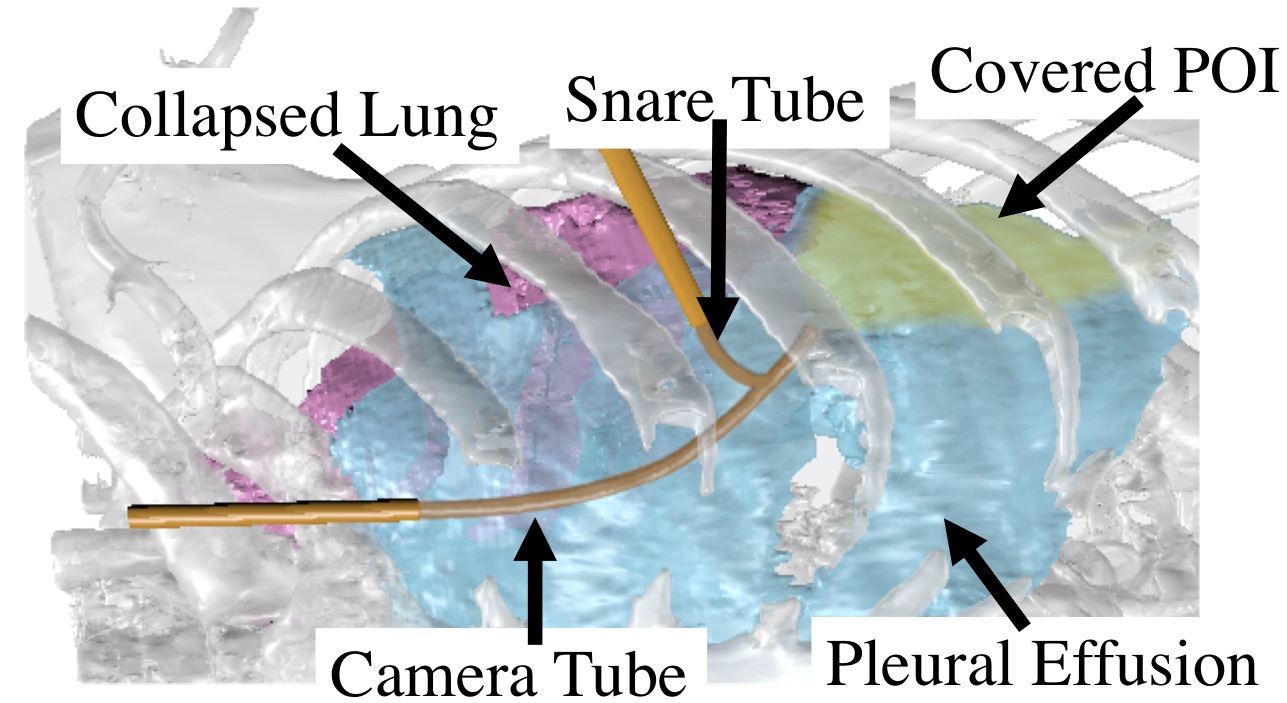}}
  \caption{Pleural effusion}
  \label{fig:effusion}
\end{subfigure}
    \caption{
        Simulation scenarios. 
            (\subref{fig:planar})~A 5-link planar manipulator (orange) inspects the boundary of a square region (blue) where rectangular obstacles (red) may block the robot and occlude the sensor. The sensor's field of view (FOV) is represented by the yellow region.
            $\S(\mathbf{q})$ are the points on the boundary in the sensor's unobstructed FOV, and are shown in purple.
            (\subref{fig:effusion})~The pleural effusion inspection scenario involves the CRISP robot (orange) inspecting the inner surface of a pleural cavity, including the POI that are covered (green) and non-covered (blue) from the current robot configuration.
    }
    \label{fig:scenarios}
\end{figure}

\begin{figure*}
\begin{subfigure}{.33\linewidth}
    \centering
     {\includegraphics[width=0.99\linewidth]{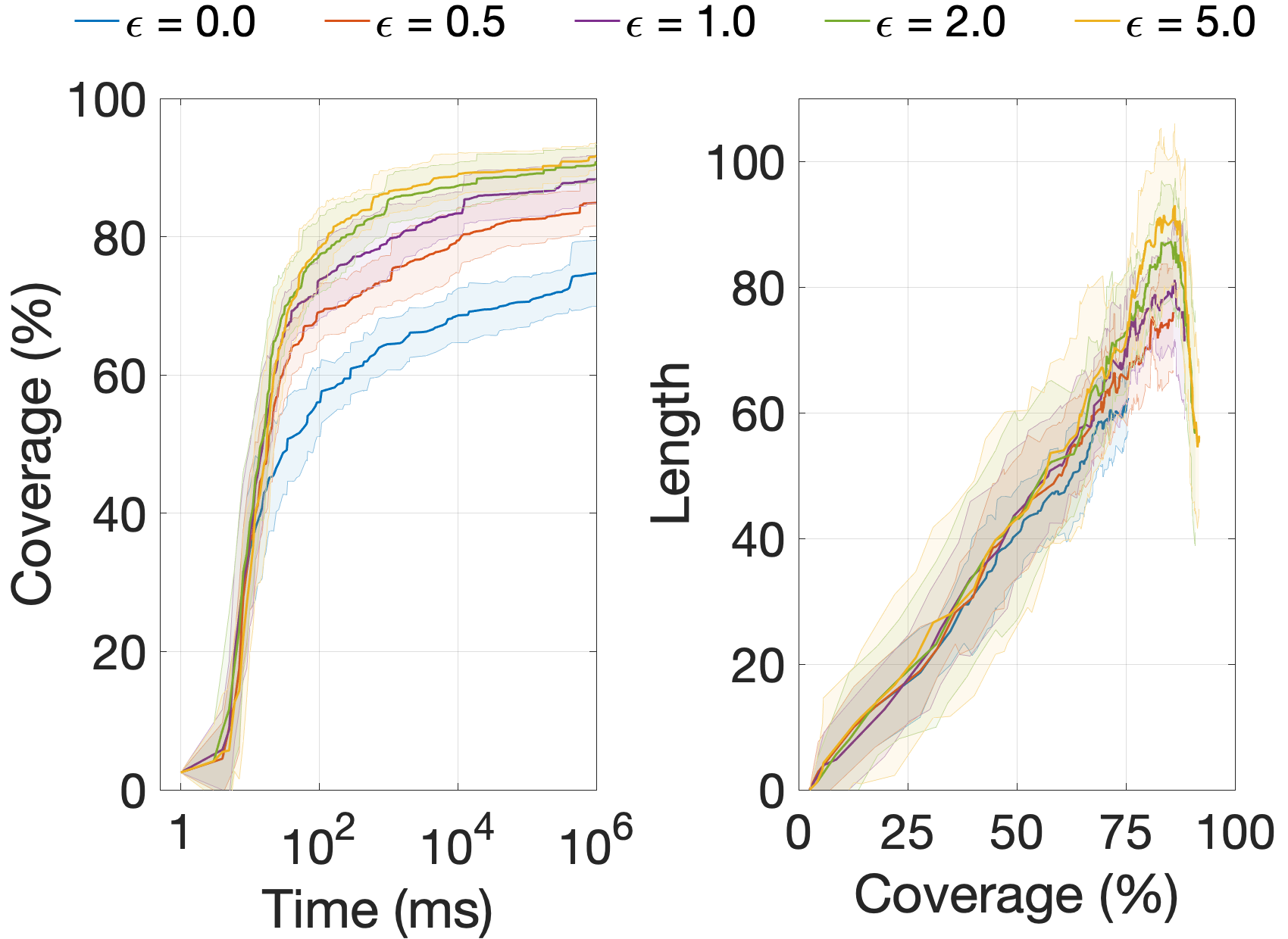}}
     \caption{}
  \label{fig:planar-fixed-p}
\end{subfigure}
\begin{subfigure}{.33\linewidth}
    \centering
     {\includegraphics[width=0.99\linewidth]{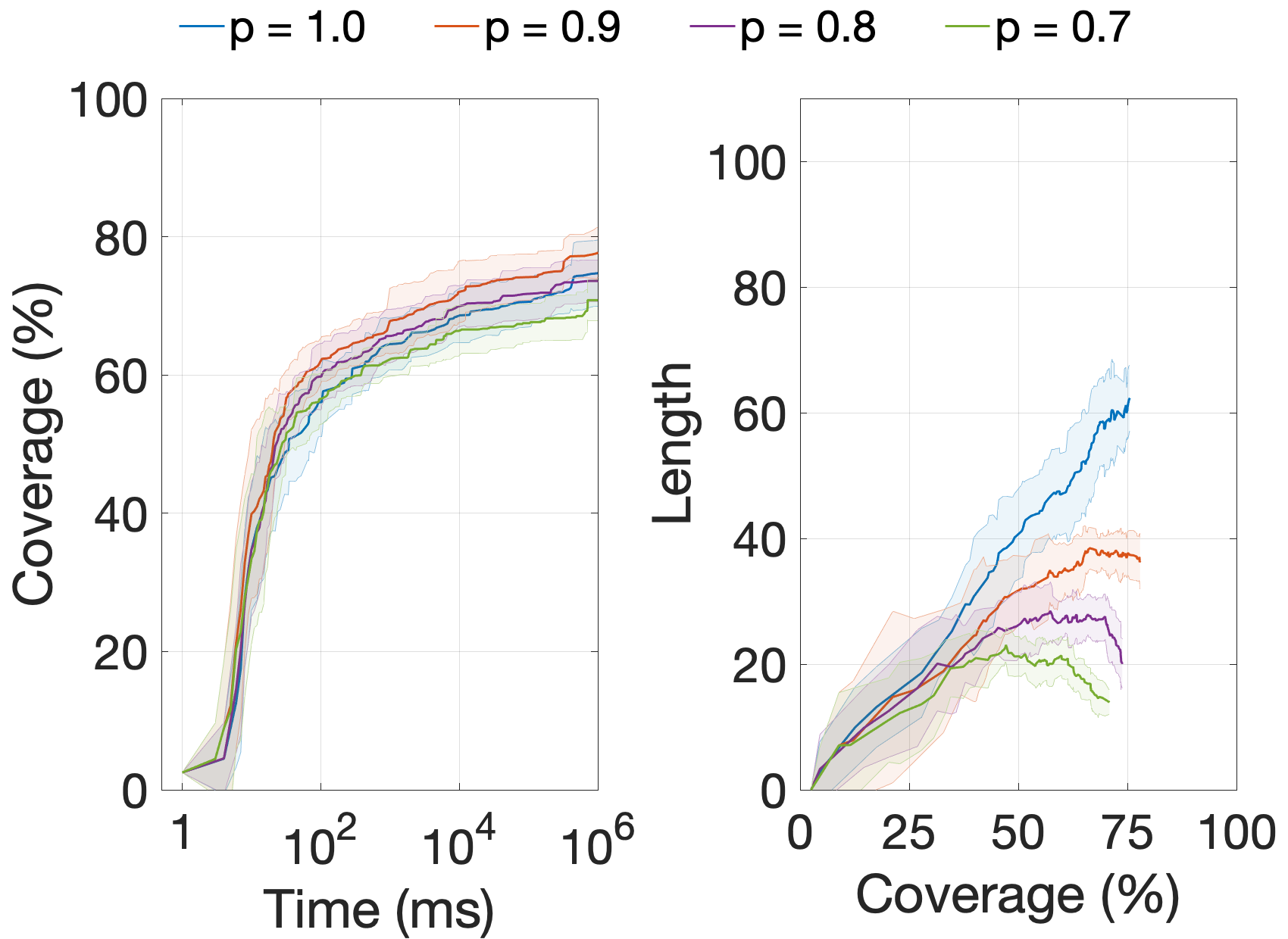}}
     \caption{}
  \label{fig:planar-fixed-e}
\end{subfigure}
\begin{subfigure}{.33\linewidth}
    \centering
     {\includegraphics[width=0.99\linewidth]{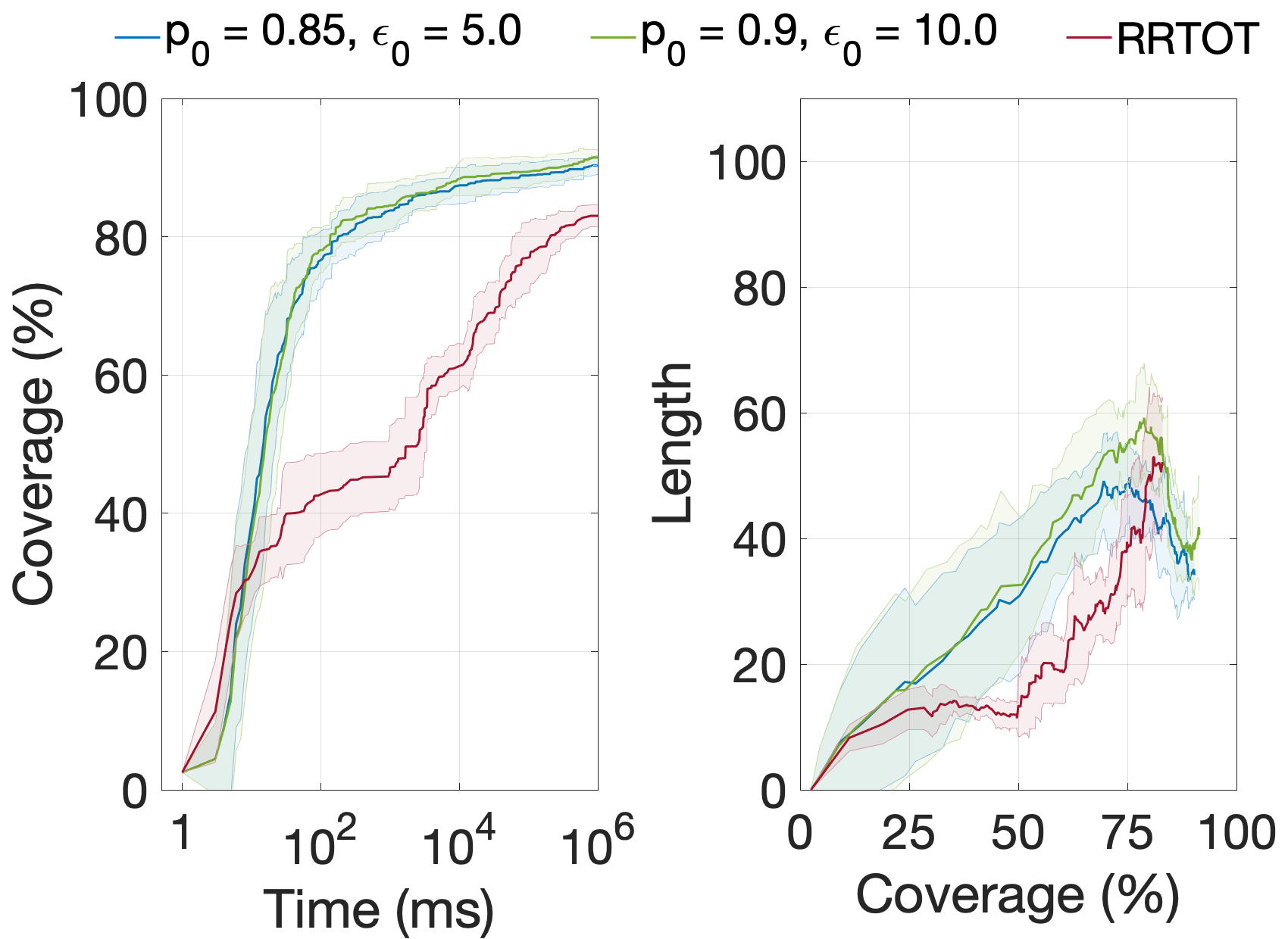}}
     \caption{}
  \label{fig:planar-varied-ep}
\end{subfigure}
    \caption{Quality of inspection paths computed for the planar manipulator.
    \protect (\subref{fig:planar-fixed-p})~\iris running with $p = 1$, $f = 0$ and varying values of $\eps$.
   \protect (\subref{fig:planar-fixed-e})~\iris running with $\eps = 0$, $f = 0$ and varying values of $p$.
   \protect (\subref{fig:planar-varied-ep})~Comparison of \iris and \rrtot. \iris running with two input parameter settings, both with $f=0.03$.
   }
\end{figure*}

We evaluated \iris on two simulated scenarios: (1) a planar manipulator inspecting the boundary of a square region (Fig.~\ref{fig:planar}) and (2) a CRISP robot inspecting the inner surface of a pleural cavity (Fig.~\ref{fig:effusion}).
All tests were run on a 3.4GHz 8-core Intel Xeon E5-1680 CPU with 64GB of RAM.

\subsection{Planar manipulator scenario}
In this scenario, depicted in Fig.~\ref{fig:planar}, we have a 5-link planar manipulator fixed at its base that is tasked with inspecting the boundary of a rectangular 2D workspace.
We start by evaluating \iris for fixed $p$ and $\eps$ and then compare it with \rrtot using our approach for dynamically reducing the approximation factors.
For every set of parameters we ran ten experiments for 1000 seconds and report the average value together with the standard deviation.

When $p = 1$ and we vary $\eps$ (Fig.~\ref{fig:planar-fixed-p}),
we can see that even small approximation factors (e.g.,~$\eps = 0.5$) allow to dramatically increase the coverage obtained as each search episode takes less time and more configurations can be added to the \rrt tree. While using $\eps = 0$ did not result in $80\%$ coverage even after 1000 seconds, this was achieved within one second for~$\eps \geq 1.0$.
This comes at the price of slightly longer inspection paths.
When $\eps=0$ and we vary $p$ (Fig.~\ref{fig:planar-fixed-e}), we get roughly the same coverage per time but at the price of much longer paths for higher values of $p$.

Following the above discussion, when reaching high coverage is the sole objective, one should use large initial values of~$p_0$ and $\eps_0$.
When we want initial solutions to also be short, one should start with smaller approximation factors.
We compared \iris with different initial approximation factors to \rrtot~\cite{BASOBS17} (Fig.~\ref{fig:planar-varied-ep}).
We can see that our approach allows to produce higher-quality paths than \rrtot.
For example, \iris obtains more than a $2450\times$ speedup when compared to \rrtot when producing  the same quality of inspection planning for the case of  roughly $83\%$ coverage and path length of $53$ units.
Final inspection paths obtained by \iris are both shorter and inspect larger portions of $\R$.

\subsection{Pleural effusion scenario for the CRISP robot}

The anatomical pleural effusion environment for this simulation scenario was obtained from a Computed Tomography (CT) scan of a real patient suffering from this condition,
and a fine discretization of the internal surface of the pleural cavity is used as the set of POI.
We also use the internal surface of the cavity as obstacles, and prohibit the robot from colliding with the pleural surface, lung, and chest wall (except at tube entry points).
Pleural effusion volumes can be geometrically complex, as the way in which the lung separates from the chest wall can be inconsistent.
This results in unseparated regions of the lung's surface that can inhibit movement and occlude the sensor from visualizing areas further in the volume.

Here we consider a CRISP robot with two tubes, where a snare tube is grasping a camera tube in order to create a parallel structure made of thin, flexible tubes.
Each tube can be independently rotated in three dimensions about its entry point into the body, and independently translated into and out of the cavity.
The system 
has 8 degrees of freedom with a configuration space of $\mathcal{SO}(3)^2\times \mathbb{R}^2$, which enables the parallel structure to move 
in a manner that enables obstacle avoidance as well as precise control of the camera's pose.

\begin{figure}
    \centering
    \includegraphics[width=0.75\linewidth]{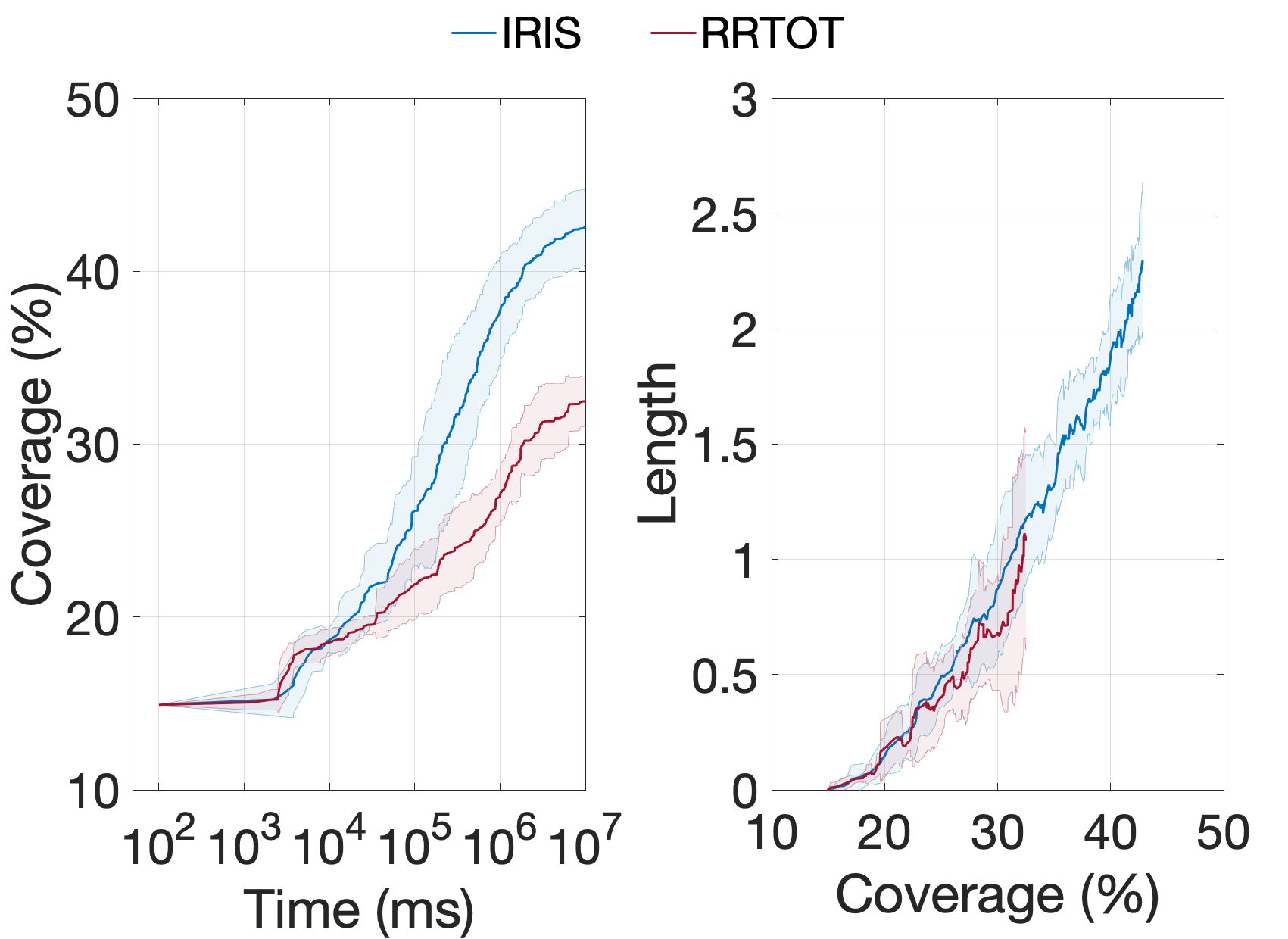}
    \caption{Comparing quality of inspection paths computed for the pleural effusion scenario.
            \iris was run with $p_0 = 0.8$,  $\eps_0=10$, and $f = 0.01$.
    }
    \label{fig:crisp_full_method}
\end{figure}

We ran \iris and \rrtot for this scenario ten different times for 10,000 seconds (Fig.~\ref{fig:crisp_full_method}).
Similar to the planar manipulator scenario, \iris allows to produce higher-quality paths than \rrtot.
For example, \iris obtains more than a $25\times$ speedup when compared to \rrtot when producing  the same quality of inspection planning for the case of  roughly $32\%$ coverage and path length of $1.2$ units.

\section{Conclusion and Future Work}
In this work we presented \iris, an algorithmic framework for computing asymptotically-optimal inspection plans.
Our key contribution is an algorithm to efficiently compute near-optimal inspection plans on graphs.
Interestingly, our problem of graph-inspection planning lies on the intersection between single and bi-criteria shortest path problems. Clearly, we are computing a shortest path on the inspection graph~$\G_\S$. However, en route we compute an approximation of the set of Pareto optimal paths to every node in the original graph~$\G$. Thus, we believe that our approach may be useful for the general problem of bicriteria optimization.

We showed \iris outperforms the prior state-of-the-art, including in a medical application in which a surgical robot inspects a tissue surface inside the body as part of a diagnostic procedure.
However, the efficiency of \iris can be further improved.
We now highlight several avenues where such improvement could be obtained.

\subsection{Dynamic updates in graph inspection planning}
\iris reruns Alg.~\ref{alg:near-optimal} every iteration which may be highly inefficient as we would like to reuse information constructed from previous search episodes.
Indeed, the general case where a graph undergoes a series of edge insertions and edge deletions and we wish to update a shortest-path algorithm is a well- studied problem referred to as the \emph{fully dynamic single source shortest-path problem}~\cite{FMN00,RR96}.
Efficient algorithms exist even when running an \astar-like search~\cite{KLF04}.
Thus, an immediate next step to improve the efficiency of our algorithm is to adapt the aforementioned algorithms to the case of near-optimal graph inspection planning.

\subsection{Balancing graph search and lazy computation}

\begin{figure}
    \centering
    \includegraphics[width=0.8\linewidth]{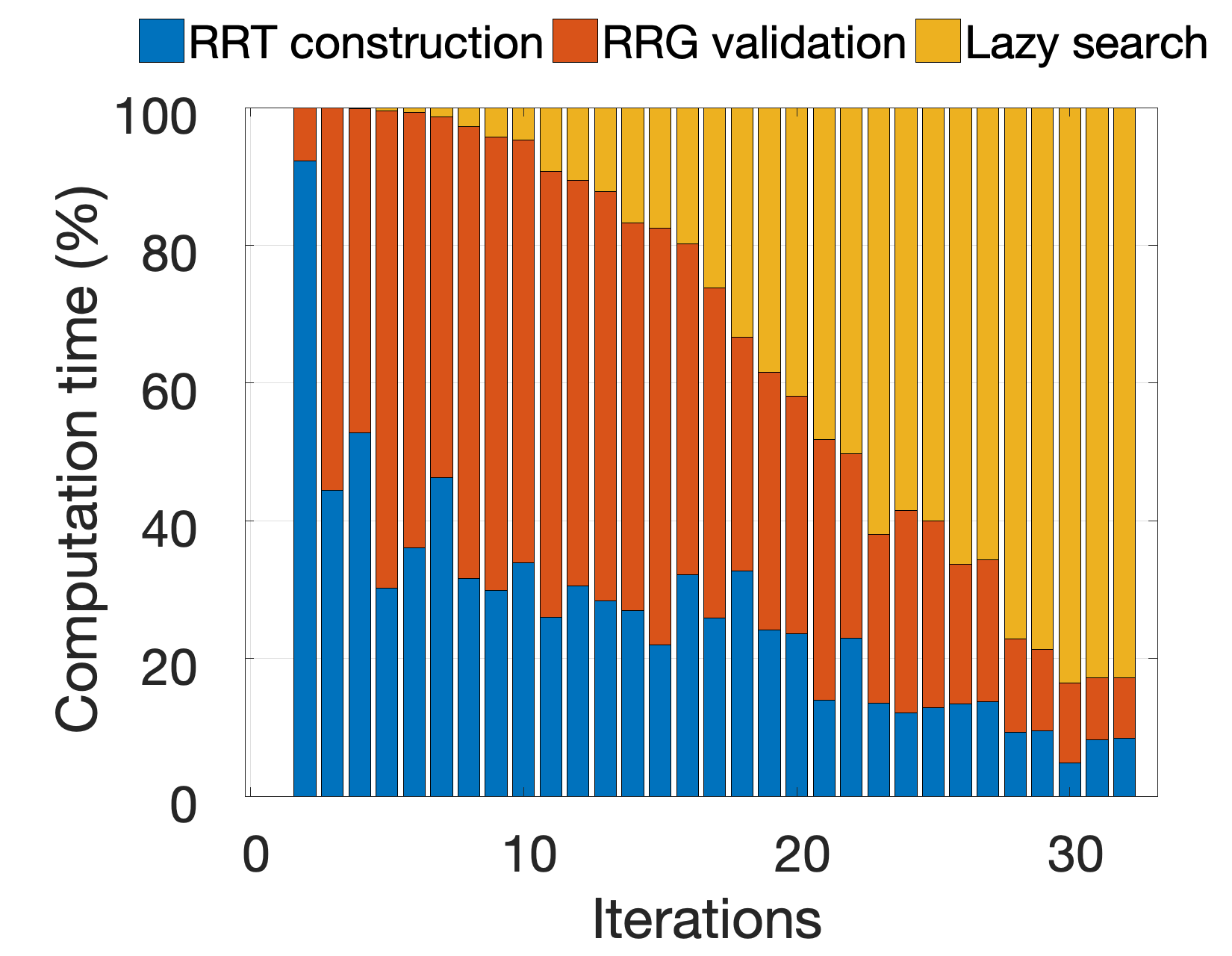}
    \caption{Time decomposition of \iris as a function of iteration number.}
    \label{fig:time_decomposition}
\end{figure}
Recall that we employ a lazy search paradigm when computing near-optimal inspection plans on the inspection graph (Sec.~\ref{subsec:implementation}).
This was done because edge evaluation is computationally complex.
However, as the number of iterations increases, search starts to dominate the overall running time of our algorithm and \emph{not} edge evaluation (see Fig.~\ref{fig:time_decomposition}).
Recently Mandalika et al.~\cite{MSS18,MCSS19}  presented an algorithm that balances edge evaluation and graph search when edges are expensive to evaluate using the notion of \emph{lazy look-ahead}.
Thus, we suggest to use their method \emph{dynamically} varying the so-called lazy look-ahead---In the initial stages of the algorithm, when search is not a bottleneck, employ a large look-ahead (which corresponds to a performing more search).
As the algorithm progress, reduce the look-ahead to account for the fact that edge-evaluation is relatively cheaper than graph search.

\subsection{Efficient sampling of configurations in \rrt construction}
Recall that in our \rrt constructions we sample configurations uniformly at random from~$\X$.
Common implementations of \rrt typically employ a \emph{goal bias} where configurations from the goal are sampled with some probability~\cite{LaValle2006_Book}.
Similarly, we suggest to bias sampling towards configurations that increase coverage. 
Namely, to configurations $\mathbf{q}$ such that $\S(\mathbf{q}) \cup \R_G \neq \emptyset$.
We suspect that the goal bias should be dynamically changed---when the inspection graph~$\G_\S$ has low coverage the bias should be high. As the overall coverage of~$\G_\S$ increase, the goal bias should be reduced to allow for shorter inspection plans.

\subsection{Employing multiple heuristics in graph inspection planning}
As the number of iterations increases, graph search dominates the running time of our algorithm.
Heuristics have been shown to be an effective tool in speeding up search algorithms and we suggest to employ recent developments from the search community to speed this part of our framework.
One such development is using \emph{multiple} heuristics to guide the search in a systematic way~\cite{ASNHL16} that has shown to be an effective tool in robot planning  algorithms~\cite{ISL18,RSL18}.

Roughly speaking, using multiple heuristics allows to encode domain knowledge without having to worry about the heuristic functions being admissible.
In our setting, we are simultaneously reasoning about inspection coverage and plan length in our graph inspection planning.
Thus, it may be beneficial to design one (or more) heuristics that account for path length and one (or more) heuristics that account for path coverage.
Then we could apply a method similar to \mhastar~\cite{ASNHL16} to combine the efforts of these heuristics.

\subsection{Adaptively updating approximation parameters}
In our work we used a simplistic approach to update the approximation parameters. These may have a dramatic effect on the quality of plans produced.
We suggest to further inspect how to update these parameters, possibly doing this in a dynamic fashion according to information obtained from previous search episodes.

\section*{Acknowledgment}
This research was supported in part by the National Institutes of Health under award R01EB024864 and the National Science Foundation under award IIS-1149965.

\end{document}